\documentclass{article} %
\usepackage[table,xcdraw,dvipsnames]{xcolor}
\usepackage{iclr2025_conference,times}

\usepackage{amsmath,amsfonts,bm}

\def\eqref#1{equation~\ref{#1}}

\def\1{\bm{1}}

\DeclareMathAlphabet{\mathsfit}{\encodingdefault}{\sfdefault}{m}{sl}
\SetMathAlphabet{\mathsfit}{bold}{\encodingdefault}{\sfdefault}{bx}{n}

\usepackage[utf8]{inputenc} %
\usepackage[T1]{fontenc}    %
\usepackage[colorlinks=true,citecolor=CadetBlue,urlcolor=CadetBlue,linkcolor=CadetBlue]{hyperref}       %
\usepackage{url}            %
\usepackage{booktabs}       %
\usepackage{amsfonts}       %
\usepackage{nicefrac}       %
\usepackage{microtype}      %
\usepackage{amsmath}        %
\usepackage{pifont}         %
\usepackage{multirow}       %
\usepackage{graphicx}       %
\usepackage{subcaption}     %
\usepackage{adjustbox}      %
\usepackage[title]{appendix}    %
\usepackage{enumitem}
\usepackage{bm}
\usepackage{diagbox}
\usepackage{titlesec}
\usepackage{wrapfig}
\iclrfinalcopy
\newcommand*\samethanks[1][\value{footnote}]{\footnotemark[#1]}

\titlespacing{\paragraph}{0pt}{1pt}{0.5em}
\titlespacing{\section}{4pt}{8pt}{0pt}
\titlespacing{\subsection}{0pt}{2pt}{0pt}

\def\poodle{\includegraphics[height=0.5cm]{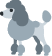}}
\def\poodlesmall{\includegraphics[height=0.35cm]{figures/1f429.eps}}

\title{\textcolor{CadetBlue}{PooDLe\poodle:} Pooled and dense self-supervised learning from naturalistic videos}

\author{Alex N. Wang$^1$\thanks{Equal contribution}\, , Christopher Hoang$^1$\samethanks[1]\, , Yuwen Xiong, Yann LeCun${}^1{}^2$, Mengye Ren$^1$\\
$^1$New York University, $^2$Meta \\
\texttt{\{anw2067, ch3451, mengye\}@nyu.edu}
\\
\url{https://agenticlearning.ai/poodle/}
}

\newcommand{\modelname}{PooDLe}
\newcommand{\wtvenice}{$\text{WT}_{\text{Venice}}$}
\newcommand{\wtall}{$\text{WT}_{\text{all}}$}

\begin{document}

\maketitle
\begin{abstract}
Self-supervised learning has driven significant progress in learning from single-subject, \emph{iconic} images.
However, there are still unanswered questions about the use of minimally-curated, naturalistic video data, which contain \emph{dense} scenes with many independent objects, imbalanced class distributions, and varying object sizes.
In this paper, we propose \modelname{}, a self-supervised learning method that combines an invariance-based objective on pooled representations with a dense SSL objective that enforces equivariance to optical flow warping.
Our results show that a unified objective applied at multiple feature scales is essential for learning effective image representations from naturalistic videos.
We validate our method with experiments on the BDD100K driving video dataset and the Walking Tours first-person video dataset, demonstrating its ability to capture spatial understanding from a dense objective and semantic understanding via a pooled representation objective.
\end{abstract}

\section{Introduction}
\looseness=-10000
Humans and other animals learn visual understanding from a continuous stream of inputs with little explicit supervision. 
Self-supervised learning (SSL)~\citep{chen2020simple,grill2020byol,chen2021simsiam,caron2021dino,bardes2021vicreg,he2022masked,assran2023ijepa,bardes2023vjepa,he2020momentummoco}
has made great strides in learning without human annotations, becoming competitive with supervised learning.
However, many methods still revolve around ImageNet~\citep{deng2009imagenet}, which is implicitly supervised through \emph{iconic} images that contain a single, clear subject and a balanced class distribution.
In contrast, naturalistic data like egocentric videos contain cluttered scenes, imbalanced classes, and objects of varying sizes, making them ill-suited for iconic methods.

\begin{figure}[t]
    \centering
    \begin{minipage}[t]{0.2145\textwidth}
        \centering
        \includegraphics[clip,trim=2.6cm 0.6cm 2.55cm 0.6cm,width=\textwidth]{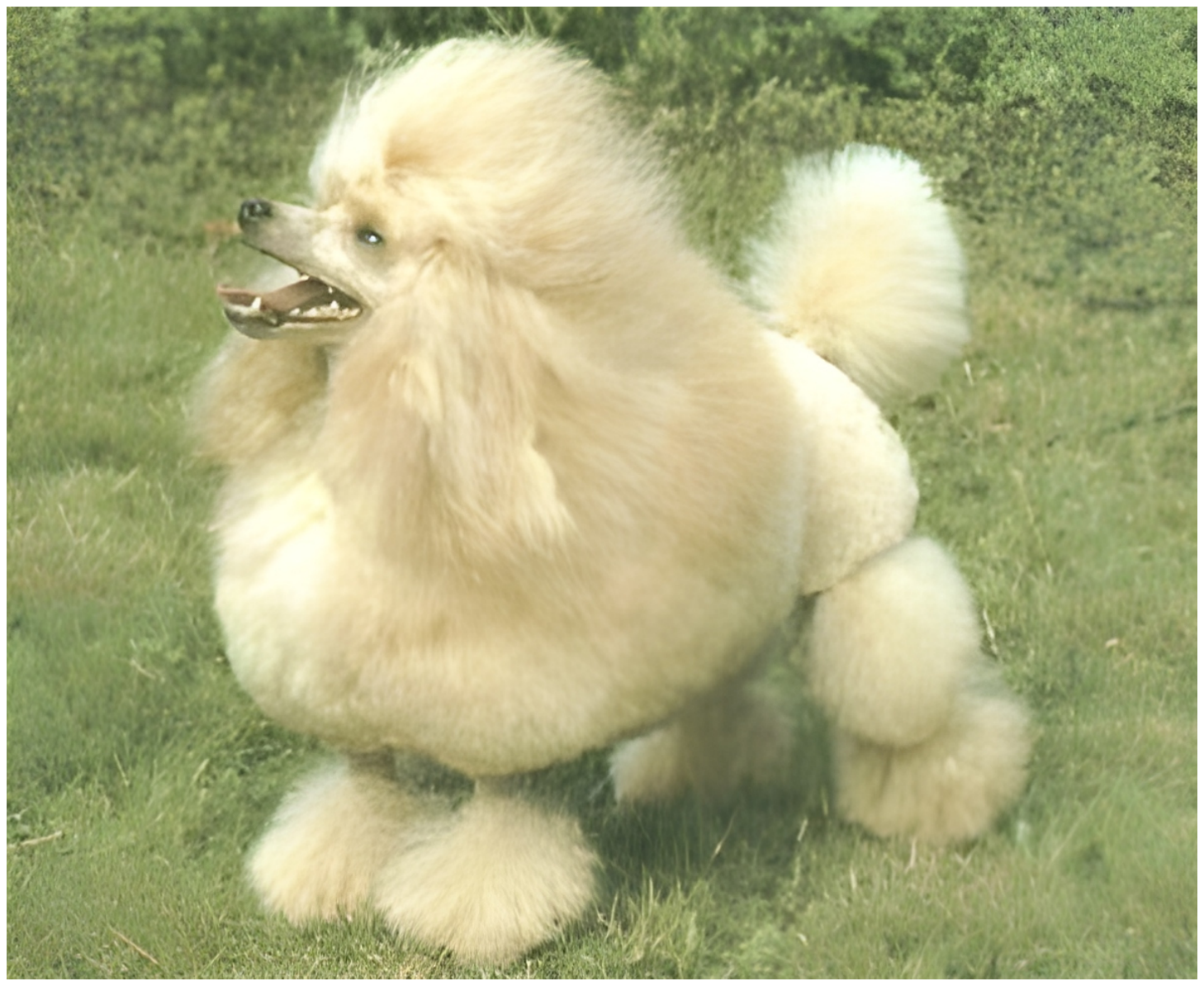}
        \subcaption{Iconic image}
        \label{fig:imagenet-image}
    \end{minipage}
    \hfill
    \begin{minipage}[t]{0.4575\textwidth}
        \centering
        \includegraphics[width=\textwidth]{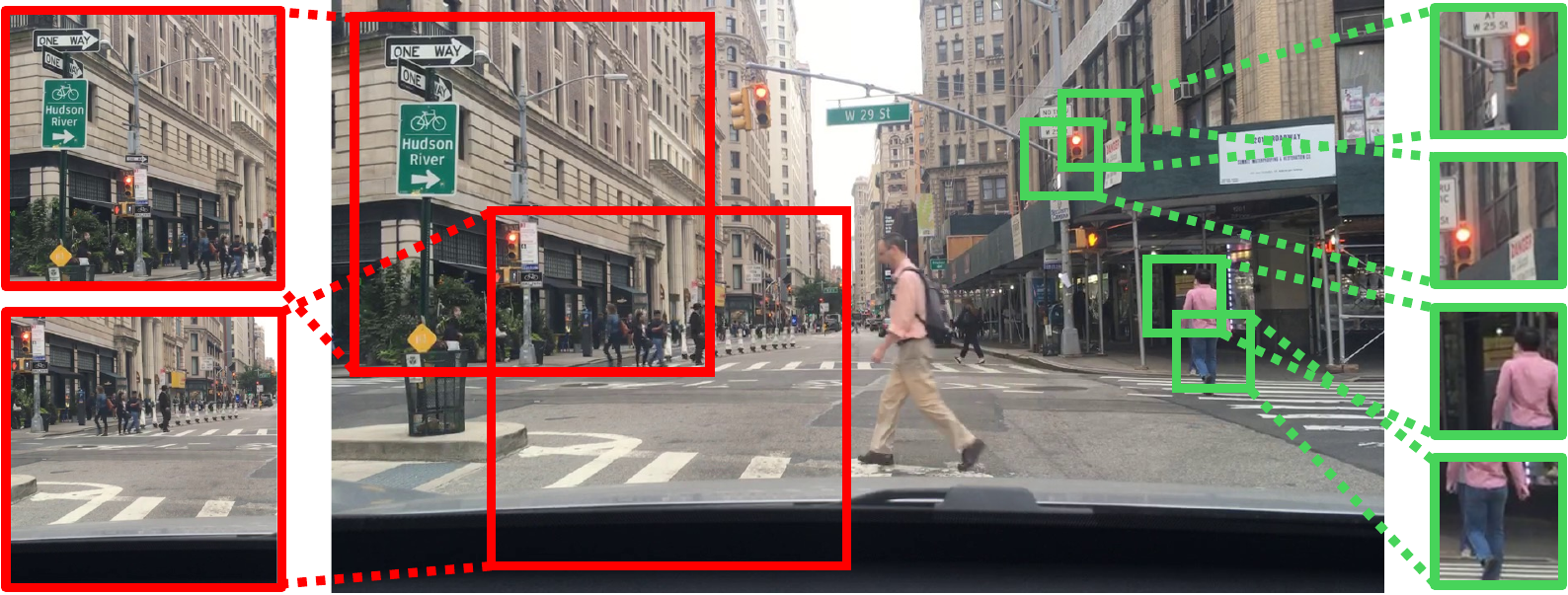}
        \subcaption{Dense scene with global crops and subcrops}
        \label{fig:bdd-dense}
    \end{minipage}
    \hfill
    \begin{minipage}[t]{0.3145\textwidth}
        \centering
        \includegraphics[width=\textwidth]{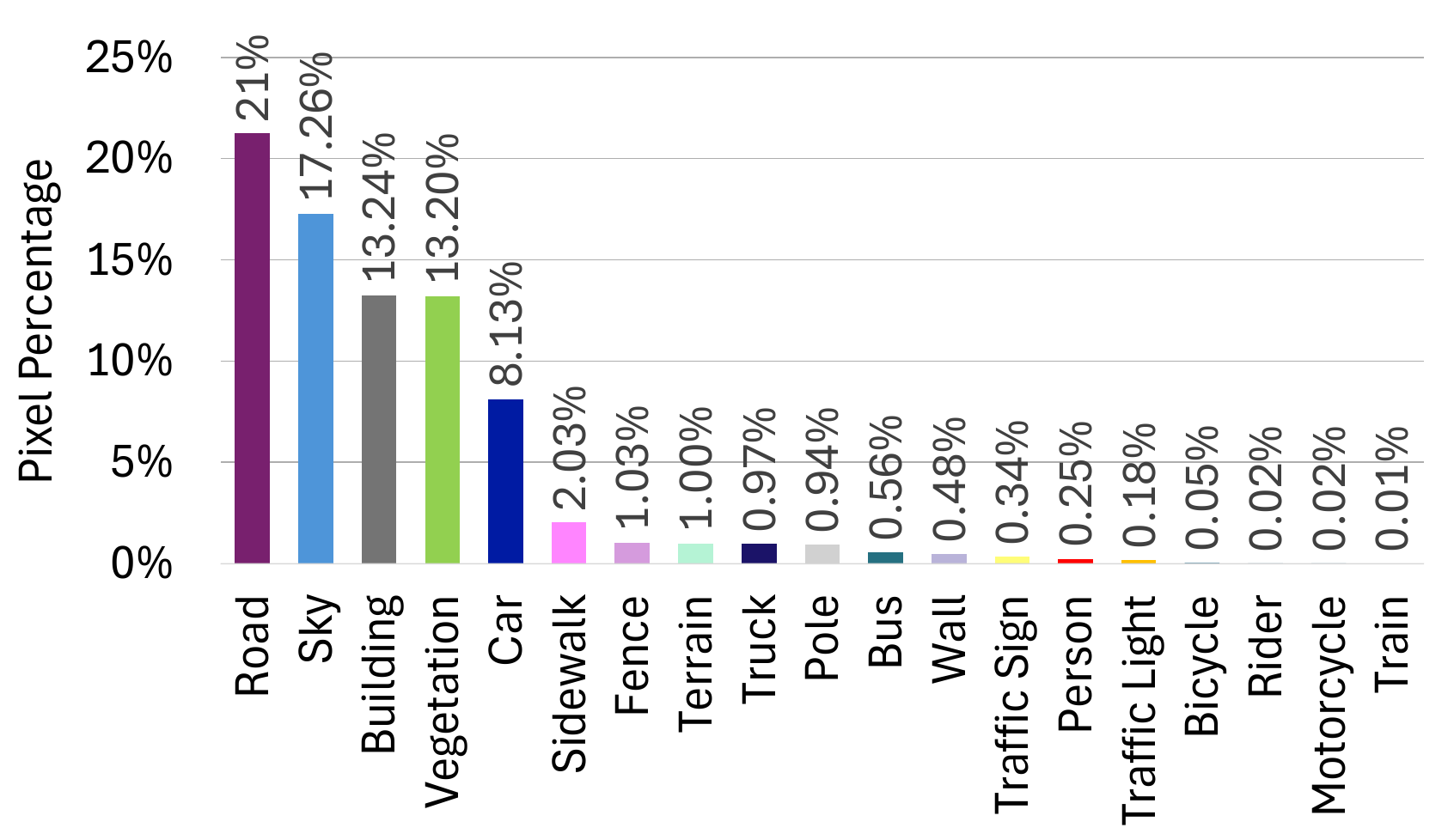}
        \subcaption{Class distribution}
        \label{fig:bdd-class-distribution}
    \end{minipage}
    \caption{\textit{Challenges of SSL on dense, naturalistic video} include training on crowded scenes with many objects of varying scale and class imbalance. (a) Iconic image from ImageNet%
    . (b) Dense scene from BDD100K%
    . Global crops (red boxes) used for iconic data contain multiple, disjoint sets of objects. Subcrops (green boxes) provide pseudo-iconic views of a single subject. (c) Naturalistic video like BDD100K have a long-tailed class distribution of class sizes and frequency; classes ordered by percentage of pixels. Example subcrops capture smaller classes like "Traffic Light" and "Person."}
    \label{fig:1challenges_of_uncurated_video}  
\vspace{-0.2in}
\end{figure}

Nevertheless, these naturalistic videos are still valuable for their information density and ease of collection, while also mimicking the real-life perspective of humans.
Unfortunately, iconic methods, which pool global image representations, may perform poorly as dense scenes produce views containing independent subjects that are semantically incompatible (Figure~\ref{fig:bdd-dense}, red boxes).
Recent works have attempted to address this weakness by introducing 1) cropping~\citep{selvaraju2021cast} or attention~\citep{venkataramanan2023imagenetdora} mechanisms to account for multiple subjects, and 2) ``dense SSL'' objectives~\citep{xiong2021selfflowe, wang2021dense} with losses defined over regions of unpooled image representations.

While dense SSL methods avoid semantic mismatches, we discover that they are susceptible to spatial imbalance where larger background classes like the sky dominate the representation, while smaller classes like pedestrians are underrepresented.
This is undesirable because smaller foreground objects should be prioritized over low-detail, repetitive background classes.
Furthermore, this can be dangerous in applications like self-driving~\citep{yu2020bdd100k} where critical objects like pedestrians occupy less than $0.3\%$ of a video frame (Figure~\ref{fig:bdd-dense}, green boxes and~\ref{fig:bdd-class-distribution}).
This contrasts with ImageNet~\citep{deng2009imagenet} training, where models can easily learn semantics from iconic images with clear, single-subject views and a balanced class distribution.
Surprisingly, dense methods like FlowE~\citep{xiong2021selfflowe} and supervised ImageNet pretraining achieve similar downstream performance while converging to very different solutions; the former prioritizes large background classes while the latter captures many small and rare classes, but with relatively poor specificity. 
Some dense SSL methods~\citep{wang2021dense, parthasarathy2023selfvito, venkataramanan2023imagenetdora} include losses that optimize a global, pooled representation to learn semantic information from dense scenes, but do not explore how to integrate these two objectives through means of architecture and augmentation strategies.

\looseness=-10000
To address the challenges of cluttered scenes and spatial imbalance,
we propose a joint \textbf{Poo}led and \textbf{D}ense \textbf{Le}arning (PooDLe) method that optimizes a dense SSL objective over full images and a pooled objective on smaller, semantically-aligned views.
The combination of objectives captures both high-level semantics and fine-grained details, effectively representing both small objects and scene-level understanding. 
Our dense objective is adopted from FlowE, using optical flow warping to align dense feature maps.
To adapt the pooled objective to dense scene data, we introduce a flow-informed cropping procedure that generates pairs of smaller ``subcrops'' with high alignment.
These subcrops serve as pseudo-iconic views of foreground objects, functionally increasing the prevalence of smaller objects (Figure~\ref{fig:1challenges_of_uncurated_video}).
Finally, we introduce a lightweight spatial decoder module (SDM) with top-down layers and UNet-like lateral connections~\citep{ronneberger2015unet} to upsample high-level semantic representations and preserve smaller objects in the dense objective.
We show that both objectives, combined with the SDM, is essential for capturing the semantics of smaller objects and achieving strong downstream task performance.

We pretrain on BDD100K, a dataset of dashcam driving videos as well as Walking~Tours, a dataset of first-person walking videos~\citep{venkataramanan2023imagenetdora}.
\modelname{} achieves state-of-the-art performance on semantic segmentation and object detection benchmarks, with a notable gain on recognizing small objects. We also introduce Walking Tours Semantic (WT-Sem) as a new in-distribution semantic segmentation evaluation for Walking Tours. 
In our ablations, we show that our joint objective formulation and the SDM are critical for success.
Finally, we study the effect of crop area, input resolution, number of subcrops and temporal stride.

\looseness=-10000
In summary, our contributions are as follows:
\begin{enumerate}[leftmargin=*]
    \itemsep -0.1em 
    \item We introduce \modelname{}, a new SSL method which overcomes the challenges of spatial imbalance and cluttered scenes by unifying a flow equivariance, dense SSL objective and a pooled objective over pseudo-iconic subcrops alongside a spatial decoder module to effectively learn from naturalistic video. 
    \modelname{} achieves state-of-the-art performance on BDD100K~\citep{yu2020bdd100k} and Cityscapes~\citep{cordts2016cityscapes} semantic segmentation and BDD object detection. 
    It also obtains the highest mIoU on ADE20K~\citep{zhou2017ade20k}, and on WT-Sem, our new, in-distribution semantic segmentation task for Walking~Tours.
    
    \item We deconstruct the BDD100K semantic segmentation task, identifying class categories by frequency and size within the dataset. We show that existing dense SSL methods and supervised ImageNet training produce different results across these categories, while PooDLe learns a balanced semantic and spatial representation to achieve strong, consistent performance.

    \item We study the effects of global and subcrop area, input resolution and temporal stride between paired frames.
    We show the importance of maintaining pixel density by adjusting crop area when training with larger resolutions for learning visual representations. We also verified that smaller subcrop areas are able to better capture smaller classes. We believe these observations will be helpful in guiding future work on dense, naturalistic data.
    
\end{enumerate}

\section{Related Work}
\paragraph{Self-supervised learning with iconic images.}
Representation learning on iconic image datasets has a long history, from denoising autoencoders~\citep{vincent2010denoisingvae} to joint embedding methods~\citep{chen2020simple,he2020momentummoco,grill2020byol,zbontar2021barlow,bardes2021vicreg,caron2021dino} to joint-embedding predictive architectures~\citep{assran2023ijepa,bardes2023mc}.
Joint embedding methods learn representation invariance to visual changes produced by augmentations using contrastive~\citep{chen2020simple, oord2018cpc}, mean squared error~\citep{grill2020byol}, or classification~\citep{caron2021dino,caron2020swav} losses between corresponding pairs, pushing SSL to new heights on ImageNet classification.
Later works extend these methods to curated, internet-scale data~\citep{oquab2023dinov2} and include other modalities like text~\citep{radford2021clip}.
Separately, MAE~\citep{he2022masked} learns via reconstruction of masked image regions.
iBOT~\citep{zhou2021ibot} combines joint embedding methods with token reconstruction to achieve impressive results on ImageNet classification.
The methods above have been primarily designed for iconic images and contain assumptions that may not transfer well to uncurated datasets, e.g. dense scenes.
Methods leveraging multi-crop~\citep{caron2020swav, caron2021dino, oquab2023dinov2, zhou2021ibot} generate small crops optimized to predict the representations of global crops for training on iconic images with little additional compute.
In contrast, our subcrop strategy yields small, aligned crops as pseudo-iconic, paired views from otherwise dense scenes. 

\paragraph{Training using dense multi-subject images.}
Following the success of SSL on ImageNet, other works seek to learn from dense, multi-subject images where augmented views may not contain corresponding subjects for invariance learning.
\citet{wang2021dense, xie2021propagate, chen2021multisiam} extend joint embedding methods by leveraging feature similarity bootstrapped from standard invariance learning to identify positive pairs across dense, unpooled feature maps.
\citet{henaff2021efficient, wang2021exploring} optimize dense losses, contrasting pixels belonging to different semantic classes; these methods require off-the-shelf segmentation modules.
\citet{ziegler2022selfleopart, guo2023multi} utilize DINO~\citep{caron2021dino} attention maps to identify training pairs, while ADCLR~\citep{zhang2023patchadclr} identifies pairs using small ``query'' crops and the patches that attend to them.
These methods advance the ability to learn from dense images with multiple objects, but still have limitations.
Some rely on learning objectives that make assumptions about iconic data, while others struggle with the spatial imbalance problem that is especially prevalent in naturalistic data.

\paragraph{Learning image representations from video data.}
Extending beyond images, other works have sought to capture the variance of objects through time by training on pairs of video frames.
\citet{gordon2020watching} adapts contrastive learning to use correlated frames as positive examples, while \citet{jabri2020spacecrw, parthasarathy2023selfvito} identify positive pairs based on high similarity in representation space.
FlowE~\citep{xiong2021selfflowe} builds on BYOL~\citep{grill2020byol} and identifies positive spatial regions between frames using off-the-shelf flow.
MC-JEPA~\citep{bardes2023mc} learns motion using video data by aligning latent representations throughout the feature pyramid while performing representation learning on ImageNet.
Most recently, DoRA~\citep{venkataramanan2023imagenetdora} proposes a new dense video dataset and extends DINO by clustering over many frames to identify and track objects for representation learning.
In the MAE paradigm, \citet{tong2022videomae,feichtenhofer2022maskedmae-st} directly reconstruct sequences of frames while~\citet{weinzaepfel2022croco,gupta2024siamese} perform reconstruction given a corresponding overlapping frame.
While \modelname{} learns a rich image representation from video data similar to these existing methods, it distinguishes itself by leveraging a unified dense and pooled objective architecture, specifically designed to tackle the challenges posed by naturalistic data.

\begin{figure}[t]
\centering
\includegraphics[width=0.98\textwidth]
{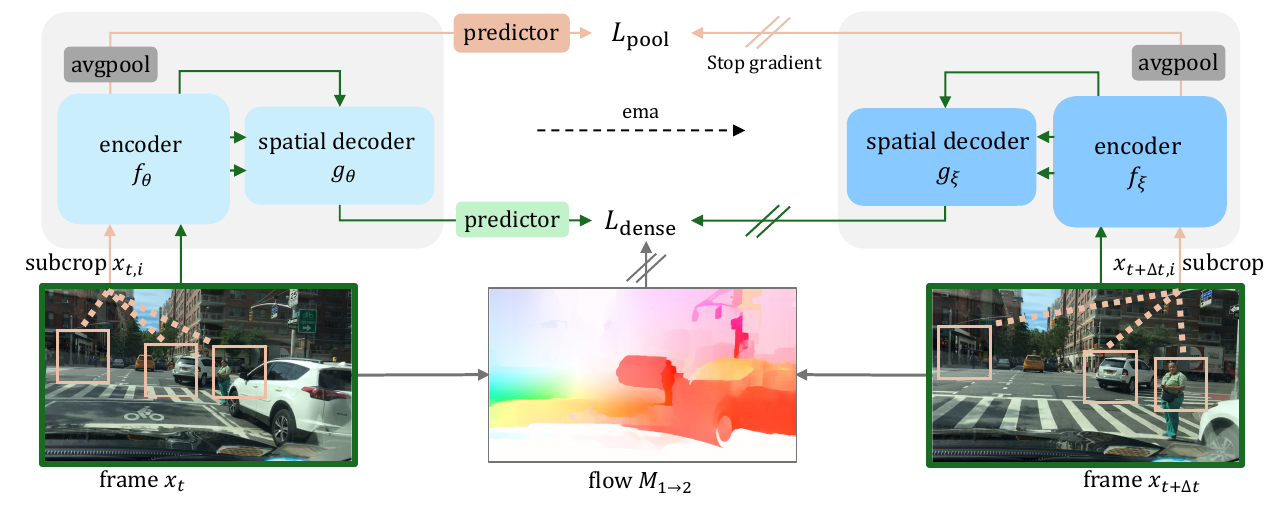}
\caption{
\modelname{}, a self-supervised learning method that combines pooled and dense objectives. \textcolor{OliveGreen}{Green} path: dense objective performing flow-equivariance learning on the output of the decoder $g(\cdot)$. \textcolor{Orange}{Orange} path: pooled objective encoding $K$ subcrops sampled with flow-informed cropping. Projector modules are not shown. Offline weights $\xi$ are the exponential moving average of online weights $\theta$. 
}
\vspace{-0.2in}
\label{fig:method1}
\end{figure}

\section{\textcolor{CadetBlue}{\modelname{}\poodlesmall:} \textcolor{CadetBlue}{Poo}led and \textcolor{CadetBlue}{D}ense \textcolor{CadetBlue}{Le}arning from naturalistic videos}

We present \modelname{}, a self-supervised method for learning visual representations using paired frames from naturalistic, first-person videos. 
\modelname{} combines two SSL objectives: a \emph{dense} objective for learning representations of dense, crowded scenes; and a \emph{pooled} objective on small subcrops sampled using flow-aware cropping augmentations. 
We also propose a lightweight spatial decoder module (SDM) that uses top-down decoder layers and UNet-like \textit{lateral} connections to earlier encoder representations to both upsample the high-level representations and resurface fine-grained details and small objects that may get lost in downsampling operations.
For a high-level overview of \modelname{}, see Figure~\ref{fig:method1}.

\paragraph{Preliminaries.}
Inputs to the model are video frame pairs $\bm{x}_t, \bm{x}_{t+\Delta t}$ with dimensions $H\times W$, and dense optical flow $M_{t\rightarrow t+\Delta t}$.
Randomly sampled augmentations $A_1$ and $A_2$ are applied to each example to create positive training pairs.
In a similar setup to BYOL, the encoder and projector are denoted as a function $\bm{p} = f(\bm{x})$ using either online weights $\theta$ or offline, EMA-updated weights $\xi$.
The predictor module $q_\theta(\cdot)$ only has online weights $\theta$.
Separate projector and predictor modules are used for the pooled and dense objectives, but are not annotated for simplicity. 
We use a ResNet-50 backbone, as well as projectors and predictors following FlowE~\citep{xiong2021selfflowe} and BYOL~\citep{grill2020byol}, that are discarded after pretraining.

\paragraph{Dense SSL with flow equivariance.}
\label{sec:method-flow-equivariance-learning}
The dense objective follows FlowE~\citep{xiong2021selfflowe} by using optical flow $M_{t \rightarrow t + \Delta t}$ to align paired feature projections $\bm{p}_{t}$ and $\bm{p}_{t + \Delta t}$. 
At a high-level, this objective minimizes differences in representation between corresponding regions.
More specifically, the inverse augmentation functions $A^{-1}$ and optical flow are used to align the representations $\bm{p}$ and after upsampling to input resolution $H \times W$, the objective is the squared error:
\begin{align}
    \mathcal{L}_\text{dense} = \frac{1}{HW} \left \lVert q_\theta(A_1^{-1}(\bm{p}_{t})) - (M_{t\rightarrow t+\Delta t} \circ A_2^{-1})(\bm{p}_{t+\Delta t})\right \rVert^2_2,
\end{align}
where normalization is applied after the predictor and flow warping.

\paragraph{Pooled objective with flow-informed subcrops.}
First, we identify $K$ pseudo-iconic subcrop pairs.
Unlike for iconic data, random crops from paired frames are unlikely to contain a common subject.
To mitigate this problem, we once again use optical flow in a \textit{flow-informed} cropping procedure to identify aligned training pairs.
For each subcrop pair, we sample a random point $(u, v)$ in the target frame $\bm{x}_{t+\Delta t}$ to serve as the crop center.
It is then warped into the earlier frame $\bm{x}_{t}$ using flow $M_{t\rightarrow t+\Delta t}$ plus random jitter $(\delta_u, \delta_t)$ for paired center $(u', v')$.
A crop is made around each center, with an area sampled from $U[s_\text{min},s_\text{max}]$ of the global crop for subcrops $\bm{x}_{t, i}$ and $\bm{x}_{t+\Delta t, i}$.

As we require both crop centers to land within the bounds of the image, subcrops tend to be center-biased~\citep{peng2022craftingcontrastivecrop} and lack diversity.
To remedy this, we employ a grid-sampling procedure for selecting the initial crop center $(u, v)$.
Each global crop $\bm{x}$ is divided into a grid with cells of side length $d_\text{grid}=\text{min}(H, W) \times \sqrt{(s_\text{min}+s_\text{max})/2}$ for a $H/d_\text{grid} \times W/d_\text{grid}$ grid.
Each cell is selected without replacement, and a center $(u, v)$ is then uniformly sampled within the cell.

After $K$ pairs $(x_{t,k}, x_{t + \Delta t, k})$ are generated, they are encoded by the backbone and the pooled objective projector. 
Unlike the dense objective, no alignment or upsampling is performed, and each projection $\bm{p}$ is averaged-pooled over its spatial dimensions before computing the loss:
\begin{align}
    \mathcal{L}_\text{pool} = \frac{1}{K}\sum_k^K \left \lVert q_\theta (\bar{\bm{p}}_{t,k}) - \bar{\bm{p}}_{t+\Delta t, k} \right \rVert^2_2,
\end{align}
where $\bar{\cdot}$ denotes average pooling over spatial dimensions followed by normalization.
Our objective has each subcrop to predict its corresponding pair, which contains the same object in a different frame.
This differs from multi-crop~\citep{caron2020swav}, where local crops predict global crops, which would be less effective for dense scenes as local crops only capture a subset of the objects in a frame.

\begin{wrapfigure}{r}{0.35\textwidth}
    \centering
    \vspace{-0.15in}
    \begin{subfigure}[b]{0.3\textwidth}
        \centering
        \includegraphics[width=\textwidth]{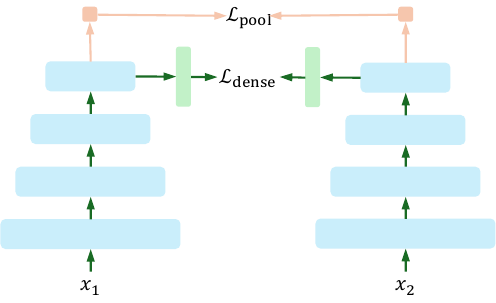}
        \caption{Baseline}
        \label{fig:sub1}
    \end{subfigure}
    \hfill
    \begin{subfigure}[b]{0.35\textwidth}
        \centering
        \includegraphics[width=\textwidth]{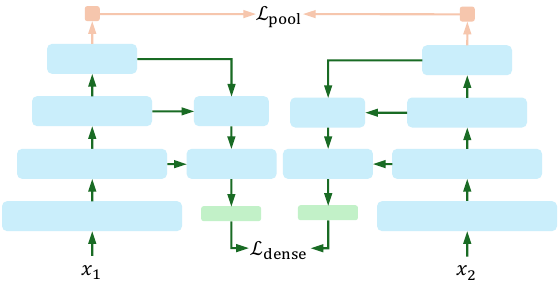}
        \caption{\modelname{} using SDM}
        \label{fig:sub3}
    \end{subfigure}
    \vspace{-0.1in}
    \caption{\textbf{a) Baseline}: both losses combined at final encoder layer; \textbf{b)\modelname{}}: 
    SDM incorporates earlier feature maps, upsampling 
    for $\mathcal{L}_\text{dense}$.}
    \label{fig:main2}
    \vspace{-0.2in}
\end{wrapfigure}

\paragraph{Spatial Decoder Module (SDM).} 

We introduce the SDM (Figure~\ref{fig:sub3}) to upsample high-level encoder features and preserve information from lower layers, particularly smaller foreground objects that may be lost during pooling operations.
Its design draws inspiration from a convolutional UNet~\citep{ronneberger2015unet} and 
FPN~\citep{lin2017fpn} and improves upon FlowE's use of dilated convolutions to replace 
pooling by efficiently maintaining high-resolution representations while reducing activations and memory usage.

The SDM utilizes decoder blocks, each consisting of an \emph{upsample} operation, a computation block of processing layers $g(\cdot)$, and a UNet-like \emph{lateral} connection.
The output of each block is computed as:
\begin{align}
\bm{z}^{l+1} = g(\text{upsample}(\bm{z}^{(l)})) + \text{lateral}(\bm{z}^{j}),
\end{align} where $\bm{z}^{(l)}$ is the representation after the $l^{\text{th}}$ encoder stage and $\bm{z}^{(j)}$ is an earlier feature map of the same spatial dimensions as $\bm{z}^{(l+1)}$.
The use of computation blocks and lateral connections is ablated in Table~\ref{table:ablations}.
Figure~\ref{fig:main2} contrasts a naive implementation that places both objectives at the top encoder level and \modelname{}, which uses the SDM to integrate the two objectives in a complementary fashion.

\section{Experiments}
\label{sec:experiments}
We pretrain \modelname{} on raw videos from BDD100K~\citep{yu2020bdd100k} and Walking~Tours~(WT)~\citep{venkataramanan2023imagenetdora} and evaluate them on semantic segmentation and object detection benchmarks.
The BDD100K pretrained model is evaluated on in-distribution tasks as well as Cityscapes~\citep{cordts2016cityscapes}, and the Walking~Tours model on ADE20K~\citep{agrawal2015learning} and our newly proposed Walking~Tours~Semantic benchmark.
We also ablate our combination of loss functions and decoder components, as well as the effects of crop area and input resolutions.

\subsection{Experiment Setup}

\paragraph{Pretraining datasets.}
    1) \textbf{BDD}~\citep{yu2020bdd100k} consists of 100,000 dashcam driving videos collected in various weather conditions and times of day from New York and the San Francisco Bay Area.
    Each video is \textasciitilde40 seconds long at 720p and 30 fps.
    We pretrain with the 70,000 videos in the training split and evaluate on the dataset's semantic segmentation and object detection tasks.
    2) \textbf{Walking~Tours (WT)}~\citep{venkataramanan2023imagenetdora} is a dataset of first-person YouTube videos of a continuous walkaround through various cities of Europe, Asia, and a wildlife safari.
    There are 10 videos, ranging from 59 minutes to 2 hours 55 minutes, at 720p and 30 fps.
    Each video contains large numbers of unique objects per frame and natural transitions in lighting and location.
    We use either the Venice video (\wtvenice{}) or all 10 videos (\wtall{}) following DoRA~\citep{venkataramanan2023imagenetdora}.

\paragraph{Technical details.} We use ResNet-50 (R50)~\citep{he2016deepresnet} as our feature encoder, with the dense projector and predictor networks following FlowE~\citep{xiong2021selfflowe} and pooled counterparts following BYOL~\citep{grill2020byol}.
For SDM, we use two decoder stages, with each consisting of a $2\times$ upsample, a ResNet Bottleneck Block~\citep{he2016deepresnet}, and a 2-layer convolutional MLP for the lateral connection.
When training on BDD, we sample two frames that are $0.5 \sim 1$ seconds apart ($\Delta t \in \{15 ... 30\}$) from each video.
We then take two large crops from the same image coordinates of area $[0.16, 0.45]$ of the original image and resize them to $512 \times 1024$ pixels before applying augmentations.
For each training epoch on WT, we divide each video into 10-second clips and randomly sample two frames $0.5$ seconds apart from each clip, and use crop area range $[0.65, 1.0]$.
For both datasets, we apply color distortion and Gaussian blurring independently to each frame following BYOL~\citep{grill2020byol}.
For the dense objective, we also apply random reversible affine transformations similar to FlowE~\citep{xiong2021selfflowe}: random scaling of 0.9--1.1$\times$ and rotation of -10--10 degrees.
For the local objective, we sample $K=6$ subcrop pairs with a crop area of $[0.05, 0.3]$ of the initial crop, resized to $192 \times 192$ for both BDD and WT.
For subcrops, random spatial jitter is applied as $\pm 10\%$ of the initial crops' height and width.

\paragraph{Baselines.} We use official implementations of DenseCL, PixPro, DINO, iBOT, and DoRA, and our own implementation of FlowE for pretraining on BDD.
We use \texttt{torchvision} for supervised ImageNet (IN1K) and weights released online for ImageNet-pretrained DINO.
We obtain weights from the authors of DoRA for iBOT, DINO-ViT, and DoRA pretrained on WT and use official implementations of DINO-R50, MAE, and PixPro for pretraining on WT.
For PixPro, we use either its FPN decoder or high-resolution crops for pretraining and report results from the best-performing setting.
We use $512\times 1024$ crops to train all R50 baselines for more accurate comparisons.

\paragraph{Evaluation.} We adopt the evaluation protocol from FlowE~\citep{xiong2021selfflowe} for BDD and Cityscapes.
We use DeepLab v1~\citep{chen2018deeplab} as the ``linear'' readout header and UperNet~\citep{xiao2018upernet} as the heavier readout head for semantic segmentation and Faster R-CNN with ResNet-C4 and Faster R-CNN with FPN~\citep{he2016faster} as the standard and heavier readout headers for object detection.
We do not include ViT object detection due to the lack of an established recipe.
For semantic segmentation on ADE20K, we perform both linear readout following BDD and UperNet finetuning as described in iBOT~\citep{zhou2021ibot}. 
We retain the SDM when evaluating \modelname{} on semantic segmentation with linear readouts but discard it when using UperNet.
We report mean intersection-over-union (mIoU), pixel-level accuracy (Acc), and mean average precision (mAP) as our evaluation metrics.
Additional details on implementation and hyperparameters are provided in Appendix~\ref{app:implementation}.

\subsection{Main Results}

\newcommand{\orangerow}{\rowcolor[HTML]{FEE9CF}}
\newcommand{\hightlightrow}{\rowcolor[HTML]{cff8ff}}

\begin{table}[t]
\caption{BDD and CityScapes semantic segmentation (SemSeg) and object detection (Det) readout evaluations. All settings are conducted with a frozen backbone. *Pretrained on BDD, initialized with supervised IN1K weights.}
\centering
\small
\resizebox{0.98\textwidth}{!}{
\begin{tabular}{@{}llll|rrrr|rr|rrrr@{}}
\toprule
& & & & \multicolumn{4}{c|}{\bf{BDD100K Sem. Seg.}} & \multicolumn{2}{c|}{\bf BDD100K Obj. Det.} & \multicolumn{4}{c}{\bf{Cityscapes Sem. Seg.}} \\
&  &  & & \multicolumn{2}{c}{\bf Linear} & \multicolumn{2}{c|}{\bf UperNet}  & \multicolumn{1}{c}{\bf Det C4} & \multicolumn{1}{c|}{\bf FPN} & \multicolumn{2}{c}{\bf Linear} & \multicolumn{2}{c}{\bf UperNet}  \\
\multirow{-2}{*}{\bf Method} & \multirow{-2}{*}{\bf Arch} & \multirow{-2}{*}{\bf Ep.} & \multirow{-2}{*}{\bf Pretrain} & \multicolumn{1}{c}{mIoU} & \multicolumn{1}{c}{Acc} & \multicolumn{1}{c}{mIoU} & \multicolumn{1}{c|}{Acc} & \multicolumn{1}{c}{mAP} & \multicolumn{1}{c|}{mAP} & \multicolumn{1}{c}{mIoU} & \multicolumn{1}{c}{Acc} & \multicolumn{1}{c}{mIoU} & \multicolumn{1}{c}{Acc} \\ \midrule
Scratch & R50 & - & - & 9.7 & 55.0 & 26.1 & 81.2 & 0.0 & 7.7 & 9.8 & 58.0 & 30.7 & 84.1\\ \midrule
DINO~\citep{caron2021dino} & ViT-S & 300 & BDD  & 29.6 & 86.8 & 41.1 & 90.1 & - & - & 35.1 & 87.9 & 51.5 & 91.9 \\
iBOT~\citep{zhou2021ibot} & ViT-S & 800 & BDD & 27.2 & 85.4 & 35.5 & 88.7 & - & - & 32.0 & 86.2 & 44.0 & 90.3 \\
DoRA~\citep{venkataramanan2023imagenetdora} & ViT-S & 200 & BDD & 33.2 & 88.1 & 43.3 & 90.7 & - & - & 37.4 & 88.7 & 50.8 & 92.0 \\
DINO~\citep{caron2021dino} & R50 & 100 & BDD & 13.1 & 64.7 & 25.6 & 80.3 & 0.3 & 11.9 & 14.9 & 69.4 & 29.2 & 81.4 \\
PixPro~\citep{xie2021propagate} & R50 & 100 & BDD & 21.8 & 80.0 & 37.3 & 88.0 & 0.7 & 18.4 & 25.5 & 81.0 & 44.3 & 89.5 \\
DenseCL~\citep{wang2021dense} & R50 & 100 & BDD & 24.2 & 84.9 & 41.8 & 90.0 & 0.7 & 20.3 & 26.6 & 85.6 & 53.2 & 91.9 \\
FlowE~\citep{xiong2021selfflowe} & R50 & 100 & BDD & 35.7 & 88.5 & 47.3 & 91.5 & 3.2 & 23.8 & 43.1 & 89.5 & 57.7 & 93.1 \\
\hightlightrow 
\modelname{} & R50 & 100 & BDD  & \textbf{39.2} & \textbf{89.2} & \textbf{49.9} & \textbf{91.8} & \textbf{4.9} & \textbf{25.2} & \textbf{47.2} & \textbf{90.2} & \textbf{60.7} & \textbf{93.5} \\ \midrule
Supervised & R50 & 600 & IN1K & 36.7 & 84.7 & \textbf{55.2} & 92.0 & 3.6 & 24.9 & 46.8 & 87.4 & 63.4 & 93.7 \\
\hightlightrow 
\modelname{} & R50 & 100  & BDD* & \textbf{44.7}  & \textbf{90.7} & 54.1  & \textbf{92.7} & \textbf{3.9} & \textbf{28.0} & \textbf{52.0} & \textbf{91.5} & \textbf{65.1} &	\textbf{94.4} \\ \bottomrule
\end{tabular}
}
\label{table:bddresults}
\end{table}

\begin{figure}[t]
    \centering
    \includegraphics[width=0.98\textwidth]{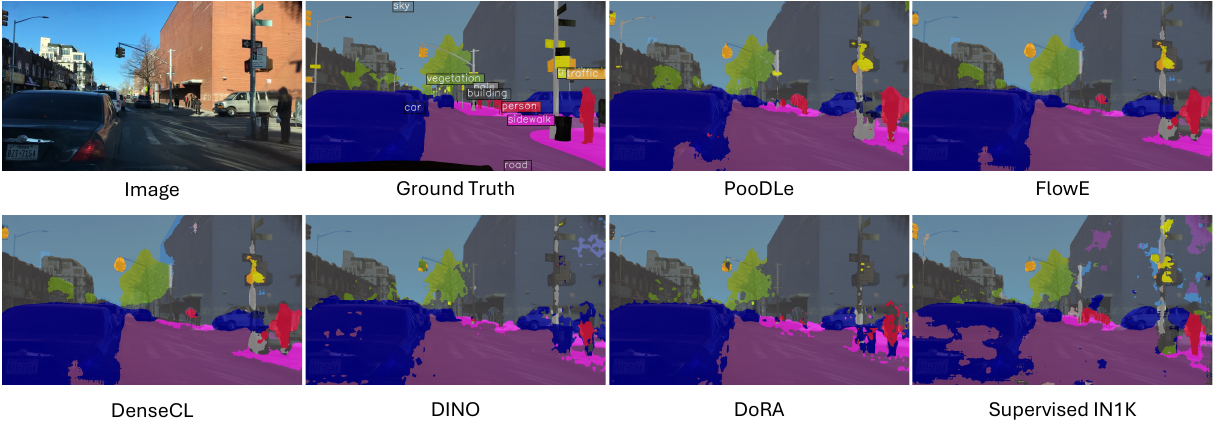}
    \caption{Visualization of BDD semantic segmentation linear readout. \modelname{} is able to identify smaller objects and generate cleaner object boundaries. 
    }
    \label{fig:bdd_visualization}
\vspace{-0.1in}
\end{figure}

\paragraph{BDD100K-pretrained models.}
We report our results on semantic segmentation and object detection on the BDD100K benchmark in Table~\ref{table:bddresults}. 
\modelname{} achieves superior performance on all readout tasks compared to prior methods, outperforming the strongest baseline FlowE by $3.5\%$ mIoU on linear and $2.6\%$ mIoU on UperNet for semantic segmentation, and $1.7\%$ mAP on C4 and $1.4\%$ mAP on FPN for object detection.
We find that that \modelname{}'s improved performance (Table~\ref{table:bdd-iou-by-class-grouping}) is attributed to better recognition of small and rare object classes.
We also evaluate the transfer of \modelname{} representations to new tasks by evaluating on the Cityscapes benchmark, where \modelname{} outperforms all baselines.
Figure~\ref{fig:bdd_visualization} shows predicted segmentation masks, and Figure~\ref{fig:big-semseg-vis} and Figure~\ref{fig:big-det-vis} show additional evaluation results.

\modelname{} also outperforms supervised IN1K pretraining, despite the latter's advantage in learning small and rare classes present in BDD100K (spatial imbalance shown in Figure~\ref{fig:1challenges_of_uncurated_video}) due to ImageNet being a class-balanced dataset with iconic views of objects.
In addition, we pretrain \modelname{} on BDD100K with weights initialized from the supervised IN1K checkpoint, improving mIoU by $8\%$ and Acc by $6\%$ over the initialization weights on linear semantic segmentation.
In Appendix~\ref{app:additional-evaluations}, we show \modelname{} remains competitive against IN1K-pretrained baselines despite being trained in the challenging naturalistic video setting.

\begin{table}[t]
\centering
\caption{ADE20K and WT-Sem semantic segmentation linear readout and finetuning evaluations. Linear readout is performed with a frozen backbone while in finetuning, backbone parameters are trainable. $^\dagger$ DINO-ViT and iBOT results are taken from DoRA~\citep{venkataramanan2023imagenetdora}. 
}
\small
\resizebox{0.99\textwidth}{!}{
\begin{tabular}{llll|rrrr|rrrr}
\toprule
& & & & \multicolumn{4}{c|}{\bf{ADE20K Sem. Seg.}} & \multicolumn{4}{c}{\bf{WT-Sem Sem. Seg.}} \\
\multirow{2}{*}{\bf Method} & \multirow{2}{*}{\bf Arch} & \multirow{2}{*}{\bf Epoch} & \multirow{2}{*}{\bf Pretrain} & \multicolumn{2}{c}{\bf SemSeg Linear}  & \multicolumn{2}{c|}{\bf Finetune} & \multicolumn{2}{c}{\bf SemSeg Linear}  & \multicolumn{2}{c}{\bf Finetune} \\ 
 & &  &  & \multicolumn{1}{c}{mIoU} & \multicolumn{1}{c}{Acc}  & mIoU & Acc & \multicolumn{1}{c}{mIoU} & \multicolumn{1}{c}{Acc} & mIoU & Acc \\ \midrule
DINO~\citep{caron2021dino} & R50  & 800 & IN1K  & 15.7 & 61.5 & 43.0 & 80.5 & 8.8 & 76.7 & 17.8 & 87.6 \\ 
DINO~\citep{caron2021dino}$^\dagger$ & ViT-S & 100 & IN1K & - & - & 33.9 & - & - & - & - & - \\
\midrule
DINO~\citep{caron2021dino} & ViT-S & 100 & \wtvenice{} & 7.8  & 57.7 & 29.2 & 74.7 & 4.6 & 73.7 & 11.0 & 83.0 \\
iBOT~\citep{zhou2021ibot}$^\dagger$ & ViT-S & 100 & \wtvenice{} & - & - & 33.9 & - & - & - & - & - \\
MAE~\citep{he2022masked}  & ViT-S & 100 & \wtvenice{} & 7.4 & 55.1 & 24.1 & 71.4 & 4.3 & 72.6 & 8.9 & 81.5 \\
DoRA~\citep{venkataramanan2023imagenetdora} & ViT-S & 100 & \wtvenice{} & 14.1 & \textbf{63.5} & 35.2 & 77.7 & 6.2 & \textbf{76.9} & 13.6 & \textbf{85.7} \\
DINO~\citep{caron2021dino} & R50 & 100 & \wtvenice{} & 6.9 & 48.2 & 35.7 & 77.4 & 4.2 & 69.0 & 12.3 & 84.7 \\
PixPro~\citep{xie2021propagate} & R50 & 100 & \wtvenice{} & 4.6 & 48.6 & 36.0 & 77.6 & 3.7 & 69.3 & 11.5 & 84.2 \\
\hightlightrow
\modelname & R50 & 20 & \wtvenice{} & \textbf{14.6}  & 59.0 & \textbf{36.6} & \textbf{77.9} & \textbf{6.4} & 75.7 & \textbf{13.7 }& 85.4 \\ \midrule
DINO~\citep{caron2021dino}$^\dagger$ & ViT-S & 100 & \wtall{} & - & - & 34.1 & - & - & - & - & - \\
MAE~\citep{he2022masked}  & ViT-S & 100 & \wtall{} & 10.6 & 60.4 & 31.4 & 75.9 & 6.6 & 77.7 & 12.7 & 85.2 \\
DoRA~\citep{venkataramanan2023imagenetdora}  & ViT-S & 100 & \wtall{} & 13.9 & \textbf{64.4} & 38.3 & 79.3 & 7.8 & 79.4 & 15.9 & \textbf{87.5} \\
\hightlightrow
\modelname & R50 & 20 & \wtall{} & \textbf{16.5} & 63.9 & \textbf{41.0} & \textbf{79.6} & \textbf{11.2} & \textbf{81.3} & \textbf{17.0} & 86.9 \\ \bottomrule
\end{tabular}
}
\label{table:wtresults}
\end{table}

\begin{table}[t]
\centering
\caption{Breakdowns of mIoU over different class groupings. Linear readout mIoU is computed over various groupings of the 19 classes in BDD semantic segmentation. *Pretrained on BDD, initialized with supervised ImageNet weights.}
\label{table:bdd-iou-by-class-grouping}
\begin{small}
\begin{tabular}{ll|r|rr|rr}
\toprule
\textbf{Method} & \textbf{Pretrain} & \textbf{All} & \textbf{Small} & \textbf{Large} & \textbf{Rare} & \textbf{Common} \\ \midrule
DINO & BDD & 29.6 & 8.4 & 42.0 & 1.0 & 42.8 \\
DenseCL & BDD & 24.2 & 1.6 & 37.4 & 0.0 & 35.4 \\
DoRA & BDD & 33.2 & 11.9 & 45.6 & 2.8 & 47.3 \\
FlowE & BDD & 35.7 & 12.2 & 49.3 & 10.7 & 47.2 \\
\hightlightrow
\modelname{} & BDD & \textbf{39.2} & \textbf{18.3} & \textbf{51.4} & \textbf{12.0} & \textbf{51.8} \\ \midrule
Supervised & IN1K & 36.7 & \textbf{27.2} & 42.2 & 16.1 & 46.2 \\
\hightlightrow
\modelname{} & BDD* & \textbf{44.7} & 25.2 & \textbf{56.1} & \textbf{17.9} & \textbf{57.1} \\
 \bottomrule
\end{tabular}
\end{small}
\vspace{-0.1in}
\end{table}

\paragraph{WT-pretrained models.}
We also train \modelname{} on \wtvenice{} and \wtall{}.
Table~\ref{table:wtresults} shows results on ADE20K~\citep{zhou2017ade20k} semantic segmentation using linear readout and finetuning following~\citep{venkataramanan2023imagenetdora}. 
Notably, when pretrained on \wtall{}, \modelname{} obtains $2.6\%$ higher mIoU than DoRA on ADE20K linear readout, and $2.7\%$ mIoU on UperNet finetuning.
\modelname{} also performs better on \wtvenice{}, with a gain of $1.4\%$ mIoU over DoRA and $0.6\%$ mIoU over PixPro on ADE20K UperNet finetuning.
We note that \modelname{} uses a smaller ResNet-50 backbone and is trained for fewer epochs than DoRA, the strongest baseline. 
Despite these differences, these results show that \modelname{} learns strong representations from naturalistic video captured in the open world.
Figure~\ref{fig:big-ade-vis} shows predicted segmentation masks for ADE20K.

\paragraph{Walking Tours Semantic benchmark.}
While ADE20K is a challenging benchmark, it contains a mixture of indoor and outdoor scenes that can be out-of-distribution from scenes in Walking~Tours. 
Therefore, we introduce Walking~Tours~Semantic (WT-Sem) to provide a more in-distribution benchmark to accompany the WT dataset. 
We find that when pretrained on \wtall{}, \modelname{} outperforms DoRA~\citep{venkataramanan2023imagenetdora} by $3.4\%$ and $1.1\%$ mIoU on linear readout and UperNet finetuning, respectively.
To generate the dataset, we use OpenSeeD~\citep{zhang2023openseed}, a strong open-vocabulary segmentation model, to generate semantic segmentation masks for all videos in \wtall{} as well as 3 new walkaround videos.
We use the Swin-L~\citep{Liu2021swin} variant of OpenSeeD finetuned on ADE20K semantic segmentation with a vocabulary of the 150 class labels from ADE20K to generate masks.
See Appendix~\ref{app:wt-sem} for visualizations and details of WT-Sem.

\begin{table}[t]
        \centering
        \captionof{table}{Ablation studies on \modelname{} components, reporting mIoU on BDD100K semantic segmentation linear readout. Rows without top-down follow FlowE~\citep{xiong2021selfflowe}, replacing pooling with dilated convolutions to maintain spatial extent. $\dagger$Flow model trained without supervised labels.}
        \small
        \resizebox{0.98\textwidth}{!}{
        \begin{tabular}{l|ccccc|r|rr|rr}
        \toprule
        \textbf{Variant} & \textbf{Dense} & \textbf{Pool} & \textbf{Top-Down} & \textbf{Lateral} & \textbf{Flow}  & \textbf{All} & \textbf{Small} & \textbf{Large} & \textbf{Rare} & \textbf{Common} \\ \midrule
        \textcolor{gray}{1} FlowE & \ding{51} & &  &  & RAFT  & 28.8 & 8.7 & 40.5 & 1.8 & 29.2  \\
        \textcolor{gray}{2} & \ding{51} & \ding{51} &  &  & RAFT & 28.9 & 7.2 & 41.6 & 2.2 & 28.7 \\
        \textcolor{gray}{3} & \ding{51}  & \ding{51} & \ding{51} &   & RAFT & 30.3 & 6.8 & 44.0 & 4.3 & 30.2 \\
        \textcolor{gray}{4} & \ding{51} & \ding{51}  &   & \ding{51} & RAFT &30.3& 10.9 & 41.7 & 2.4 & 31.1 \\
        \textcolor{gray}{5} & \ding{51} &  & \ding{51} & \ding{51} & RAFT & 31.8 & 12.8 & 42.8 & 8.3 & 31.7 \\
        \hightlightrow \textcolor{gray}{6}  \modelname{}$\dagger$ & \ding{51}  & \ding{51}  & \ding{51} & \ding{51} & UFlow & 33.7 & 14.1 & 45.1 & 8.9 & 33.8 \\
        \hightlightrow \textcolor{gray}{7}  \modelname{} & 
        \ding{51} & \ding{51} & \ding{51} & \ding{51} & RAFT & \textbf{34.2} & \textbf{15.0} & \textbf{45.5} & \textbf{9.0} & \textbf{34.5} \\ \bottomrule
        \end{tabular}
        }
        \label{table:ablations}
\end{table}

\begin{figure}[t]
\centering
\begin{minipage}{0.32\textwidth}
    \centering
    \setlength{\tabcolsep}{4pt}
    \small
    \begin{tabular}{c|ccc} 
    \toprule
    & \multicolumn{3}{c}{\textbf{classes}} \\
    \textbf{Subcrop area} & \textbf{small} & \textbf{large} & \textbf{all} \\ \midrule
    0.04 - 0.18 & \textbf{16.0}           & \textbf{41.6}           & \textbf{34.6}          \\ 
    0.18 - 0.36 & 14.2           & 41.2           & 33.7          \\ 
    0.36 - 0.54 & 13.6           & 39.5           & 32.4          \\ 
    0.54 - 1.00 & 12.3           & 38.8           & 31.5          \\ \bottomrule
    \end{tabular}
    \captionof{table}{Choice of \textbf{subcrop area} on small, large and all classes.}
    \label{tab:subcroparea}
\end{minipage}
\hfill
\begin{minipage}{0.64\textwidth}
    \begin{subfigure}{0.49\textwidth}
        \includegraphics[clip,trim=1.2cm 1.1cm 1.12cm 1.15cm,width=\textwidth]{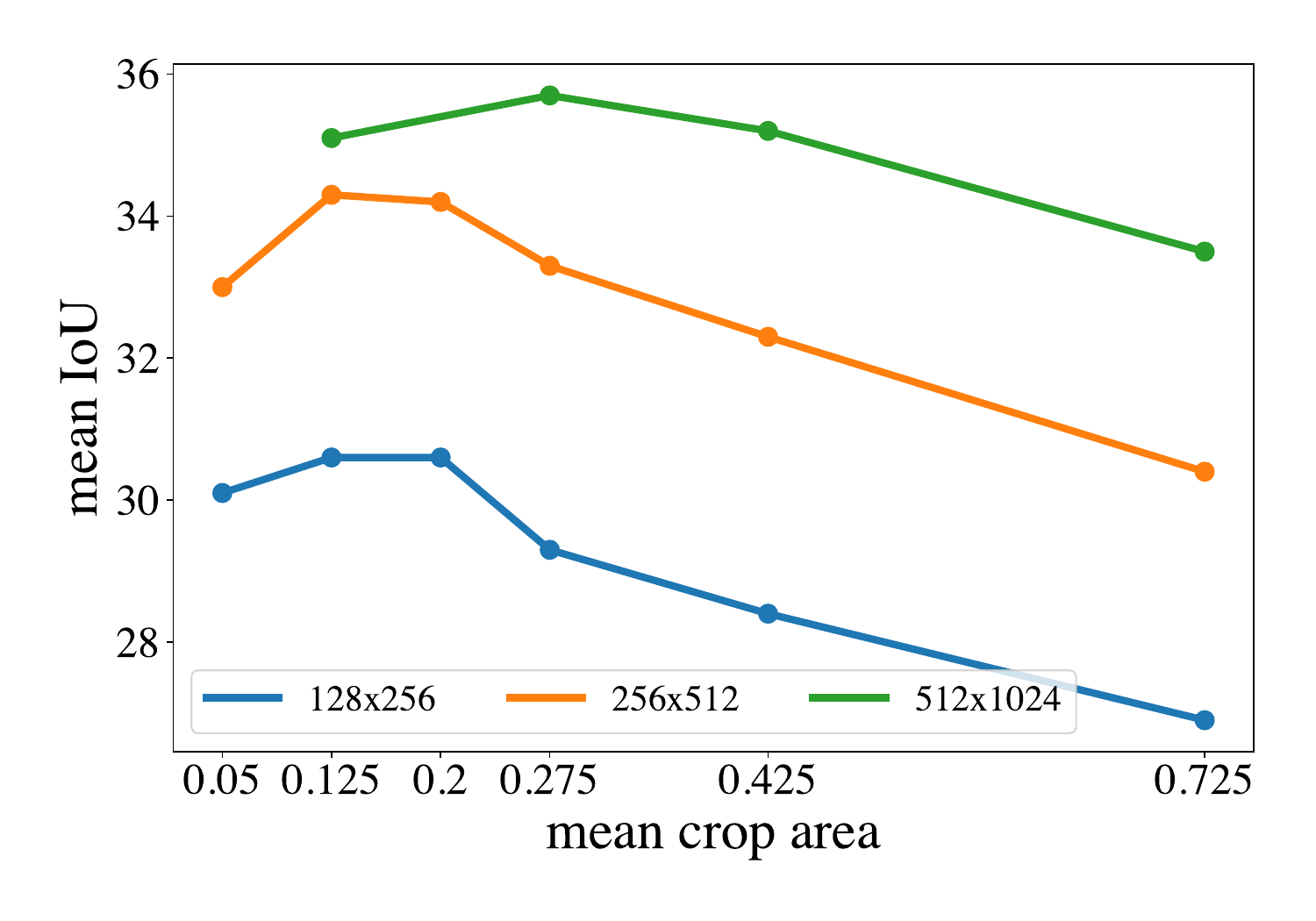}
    \end{subfigure}
    \begin{subfigure}{0.49\textwidth}
        \includegraphics[clip,trim=1.2cm 1.1cm 1.2cm 1.15cm,width=\textwidth]{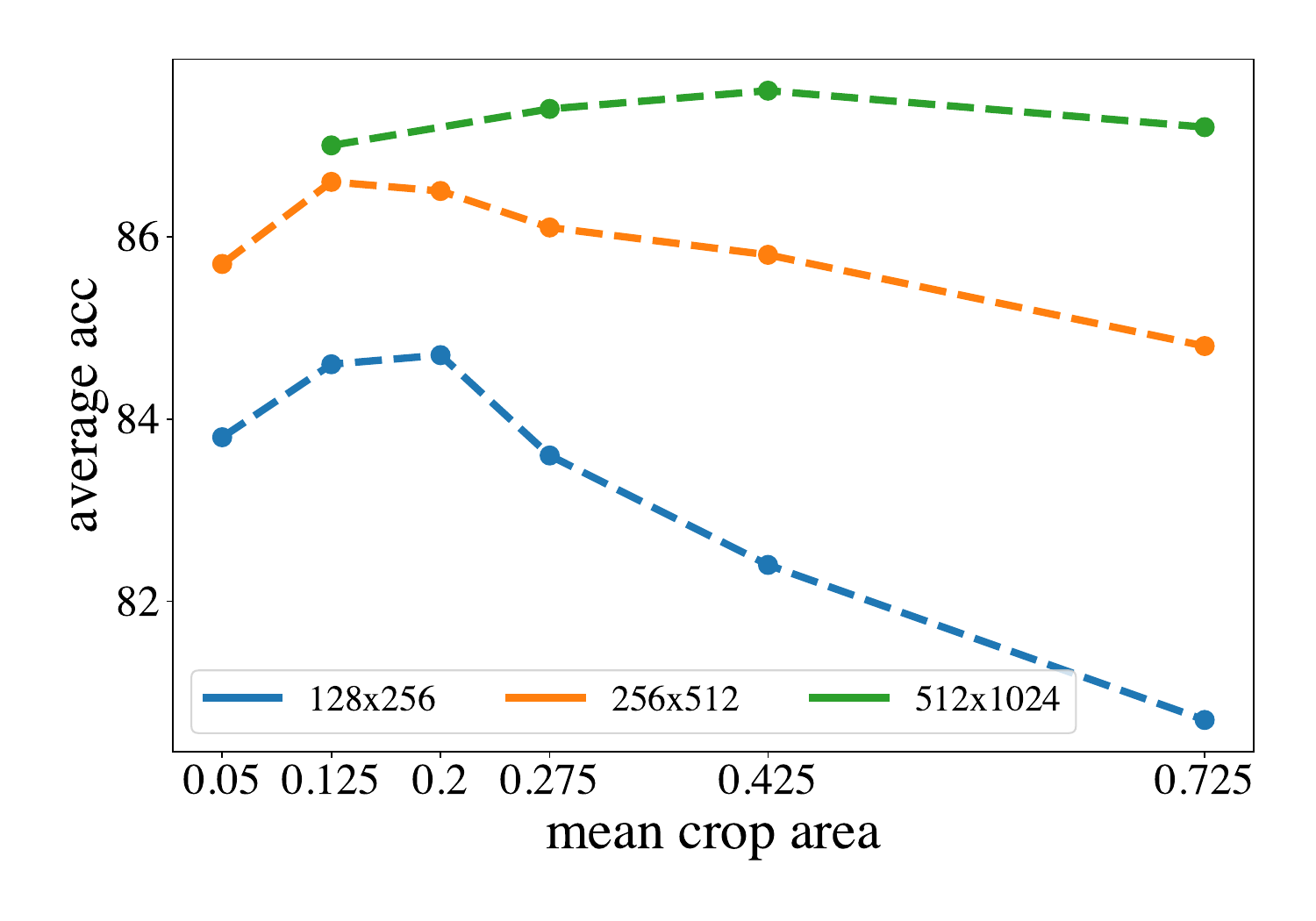}
    \end{subfigure}
    \vspace*{-2mm}
    \caption{Varying \textbf{input resolution} and \textbf{global crop area} as a fraction of the full frame. Large resolutions prefer larger crops.
    }
    \label{fig:vary_resolution_and_croparea}
\end{minipage}
\vspace{-0.1in}
\end{figure}

\paragraph{Class-based performance and IN1K initialization.}
Naturalistic videos have imbalanced class representation and object sizes (Figure~\ref{fig:bdd-class-distribution}; e.g., ``road'' occupies $21\%$ of pixels while ``bicycle'' only occupies $0.05\%$ of pixels).
Capturing information on these underrepresented classes is very challenging.
To further demonstrate this phenomenon, we categorize BDD classes as ``small'' if they occupy $<1\%$ of pixels and ``large'' for those that occupy $>1\%$.
Separately, we define ``rare'' as classes that appear in $<20\%$ of images and ``common'' as those that appear in $>20\%$.
Table~\ref{table:bdd-iou-by-class-grouping} shows linear readout mIoU for different class groupings, highlighting the impact of class and spatial imbalance.
Full class-level statistics and designations are in Appendix~\ref{app:per-class-evaluations}.
We observe that FlowE performs well on large and common classes due to its dense loss, but struggles on small or rare classes.
Meanwhile, supervised IN1K, benefiting from balanced pretraining data, effectively learns about smaller classes.
\modelname{}, with its unified objectives and spatial decoder module, significantly outperforms other BDD-pretrained models across all class groupings, particularly on small and rare classes.
\modelname{}, initialized from supervised IN1K weights, significantly improves upon supervised IN1K on large classes, from $42.2\%$ to $56.1\%$, due to the dense objective, while remaining competitive with supervised IN1K on small classes.

\subsection{Ablation studies}
\label{sec:method-ablations}
Table~\ref{table:ablations} shows ablation experiments, testing each of our contributions beginning from FlowE.
Models trained without the decoder use dilated convolutions in place of pooling operations, as in FlowE~\citep{xiong2021selfflowe}. 
Figure~\ref{fig:main2} shows how the dense and pooled objectives are composed with and without the decoder.
For the ablations, models are trained for $40$ epochs on BDD and use a reduced $256 \times 512$ resolution and $[0.04, 0.11]$ area for the initial crops; we evaluate on BDD semantic segmentation using linear readout.

We observe that adding $\mathcal{L}_\text{pool}$ alone has little benefit (row 2) and including either the decoder as a spatial upsampler (row 3) or only the UNet-style lateral connections (row 4) also does not yield much benefit.
Row 5 achieves +3\% mIoU, showing that the top-down decoder is only effective when combined with the lateral connections for the full SDM, suggesting that preserving high-resolution information as well as including some capacity for feature processing are both important.
However, when re-adding the pooled loss in addition to the decoder \textit{with} lateral connections (row 7), we see a substantial 5.4\% mIoU improvement.
While the dense objective benefits from the full SDM, it has an even greater synergistic effect with the pooled loss.
This may be because the pooled objective with subcrops can effectively learn about small objects while the full decoder helps propagate the semantic representations through to the dense loss.

\begin{figure}[t]
    \centering
    \begin{minipage}[t]{0.68\textwidth}
        \includegraphics[clip,trim=0cm 0.35cm 0cm 0cm,width=\textwidth]{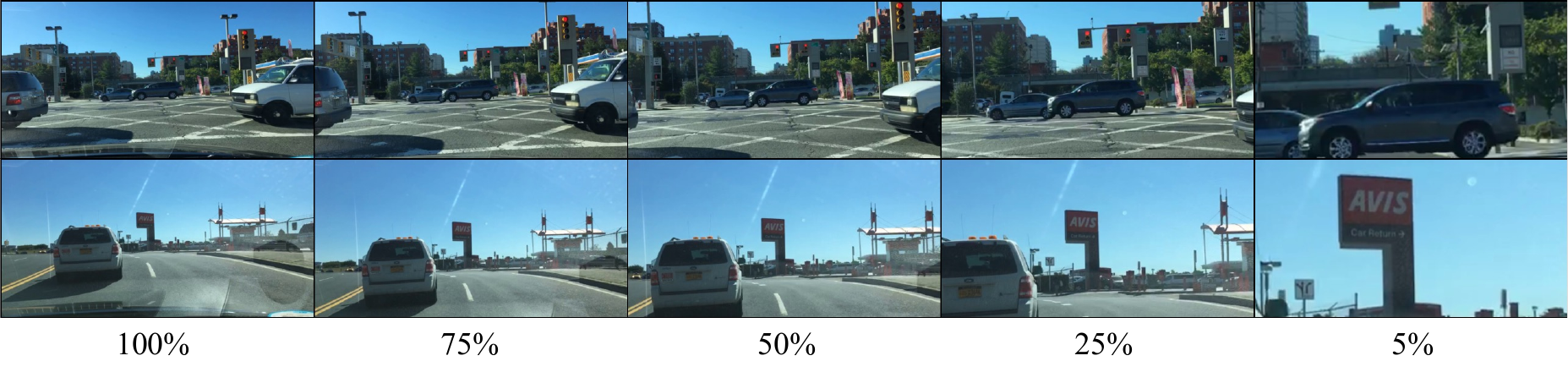}
        \captionof{figure}{Varying initial crop area as a percent of the full frame. As crop area decreases, the views transition from global views of a dense scene to pseudo-iconic views, sometimes depicting singular subjects.
        }
        \label{fig:global-to-iconic-views}
    \end{minipage}
    \hfill
    \begin{minipage}[t]{0.3\textwidth}
        \centering
        \includegraphics[clip,trim=0.2cm 1cm 0.2cm 0.2cm,width=\textwidth]{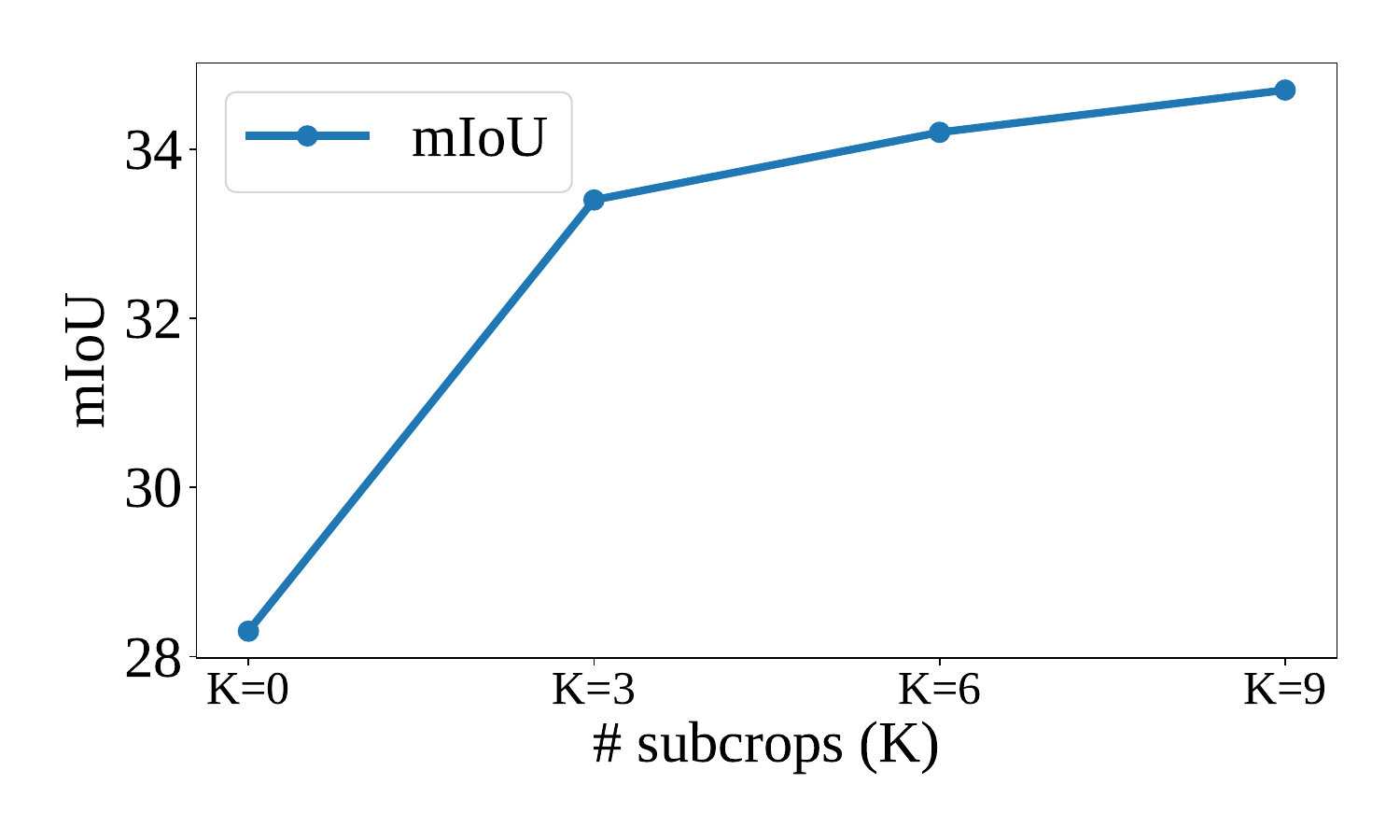}
        \captionof{figure}{mIoU when varying number of subcrops $K$.}
        \label{fig:vary-subcrops}
    \end{minipage}
\vspace{-0.1in}
\end{figure}

\looseness=-10000 We also demonstrate that \modelname{} is able to perform well even with self-supervised flow.
We train UFlow~\citep{jonschkowski2020uflow} model on KITTI~\citep{geiger2013kitti} and finetune it on BDD resulting in only a 0.5\% mIoU loss compared to pretraining with RAFT.
See Appendix~\ref{app:implementation} and~\ref{app:selfsup-flow} for details and visualizations.

\subsection{Spatial and temporal cropping in self-supervised video learning}
In this section, we study the effect of frame intervals and image cropping parameters used in data augmentation.
Without a 1:1 image-to-concept relationship like in iconic data, the visible area of each frame can greatly affect representation learning.
To study this, we perform 4 experiments varying: (1) subcrop area, (2) global crop area, (3) number of subcrops, (4) temporal stride $\Delta t$.
Crop and subcrop area refer to the fraction of the frame taken during random-resized cropping.
Figure~\ref{fig:global-to-iconic-views} depicts how crops transition from global to pseudo-iconic with decreasing crop area.
Training recipe follows the ablations in section~\ref{sec:method-ablations} and results should be compared to row 7 in Table~\ref{table:ablations}.

\begin{figure}[t]
    \centering
    \begin{minipage}{0.65\textwidth}
        \includegraphics[clip,trim=0cm 0.35cm 0cm 0cm,width=\textwidth]{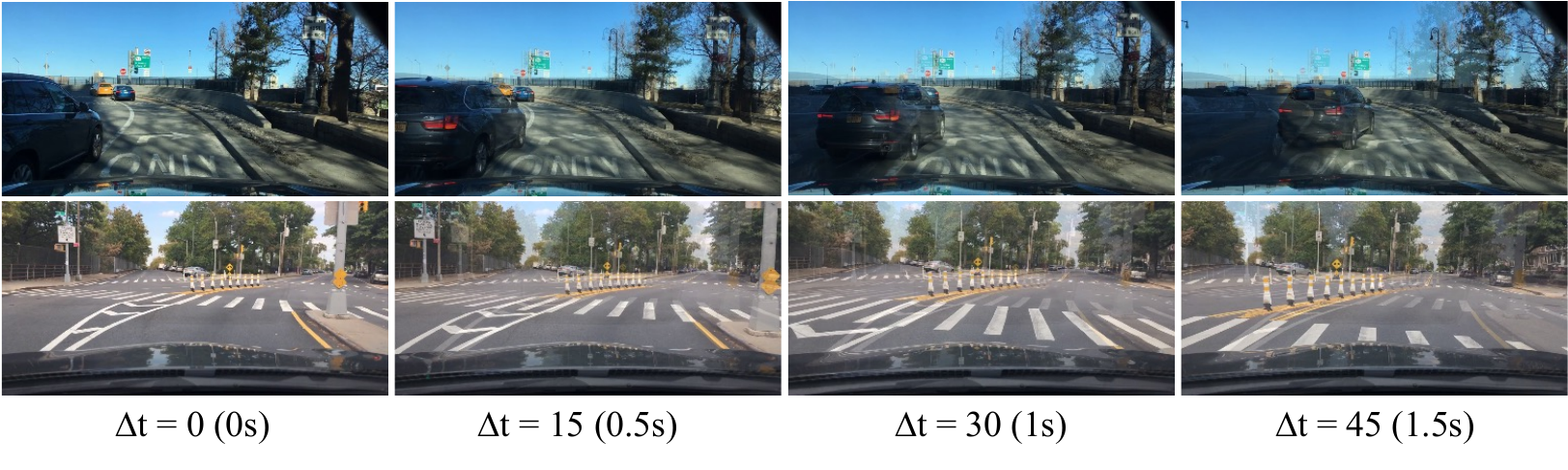}
        \caption{Overlaid frames with varying $\Delta t$.
        }
        \label{fig:delta_t-vis}
    \end{minipage}
    \hfill
    \begin{minipage}{0.34\textwidth}
        \includegraphics[clip,trim=1.2cm 1.2cm 1.2cm 1.15cm,width=\textwidth]{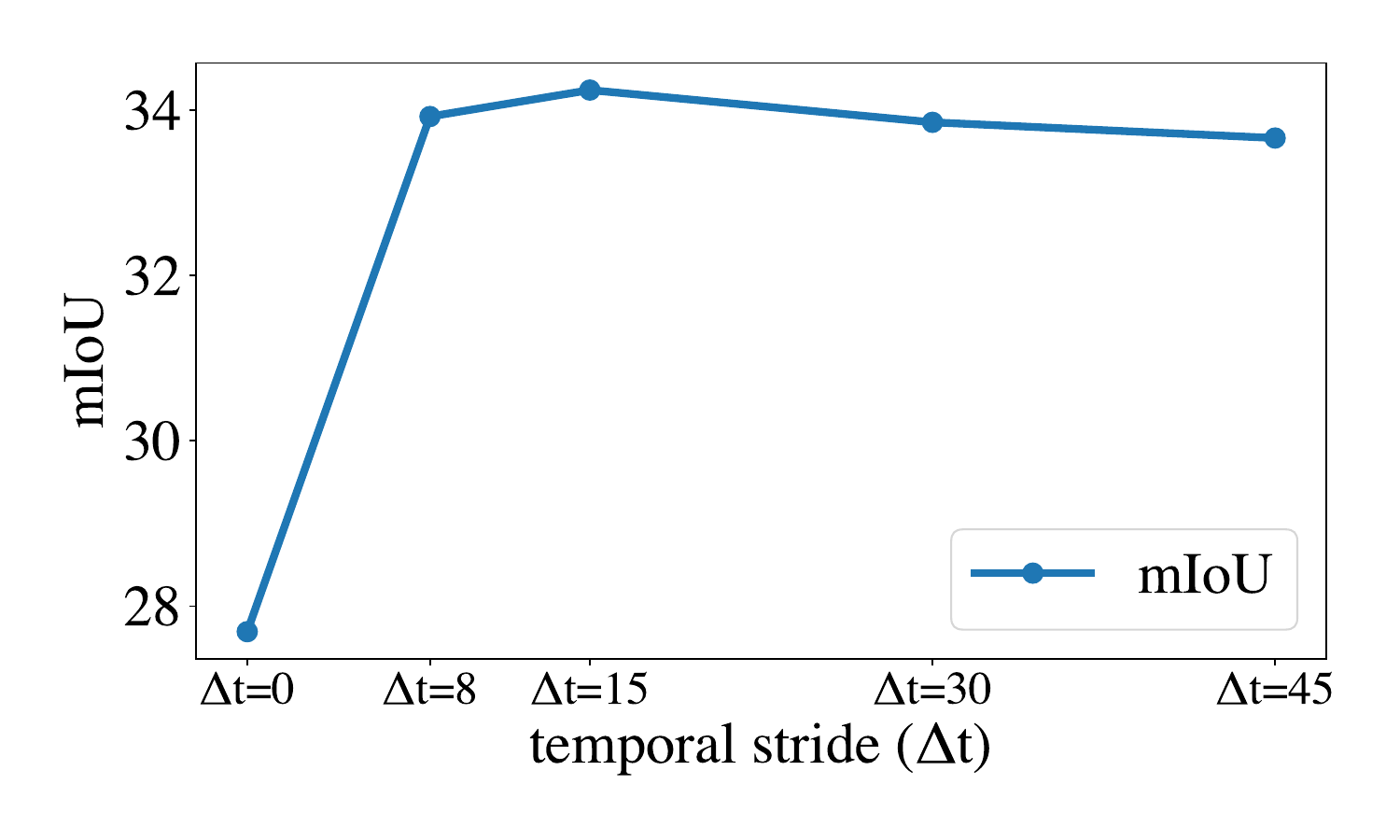}
        \caption{mIoU when varying $\Delta t$.}
        \label{fig:delta_t-sweep}
    \end{minipage}
    \vspace{-0.2in}
\end{figure}

\paragraph{Varying subcrop area.}
First, we study how subcrop area affects our learned representations.
We train 4 \modelname{}s using different subcrop ranges and a fixed global crop area of [0.125, 0.25] at resolution $256 \times 512$.
Results are shown in Table~\ref{tab:subcroparea} for all classes, as well as small and large class subgroupings.
We observe that larger subcrop areas result in worse performance, with a larger relative drop for smaller classes. 
This is likely because larger crops contain multiple, smaller objects which breaks the pseudo-iconic assumption and produces false invariances.

\paragraph{Varying global crop area}
Next, we vary both the global crop area from the raw video frame along with the input resolution to study their effects on self-supervised pretraining.
We select 3 different resolutions, and sample the crop area from a Gaussian truncated to $[0, 1]$, with varying mean and $\sigma=0.1$. 
All three resolutions are trained with $\mu=0.125, 0.275, 0.425, 0.725$ with the 2 smaller resolutions also trained at $\mu=0.05, 0.20$ for higher granularity. 
Our results in Figure~\ref{fig:vary_resolution_and_croparea} show that larger crop areas and higher input resolutions, together, are important for maximizing performance.
The largest model ($512\times 1024$) produces the best results and peaks at $\mu=0.425$ while the other two peak at smaller crop areas.
The $256\times 512$ model degrades more slowly in performance as crop area increases in comparison to the $128\times 256$ model.

\paragraph{Varying number of subcrops.}
We also study how varying the number of subcrops affects performance.
We train 4 \modelname{}s using $K=0, 3, 6, 9$ subcrops on BDD100K and evaluate using linear readout, with results shown in Figure~\ref{fig:vary-subcrops}.
Using 3 subcrops gives an initially large performance jump, and using additional subcrops provides more modest gains.
We decide to use $K=6$ as our default option to balance between performance and computational efficiency.

\paragraph{Effect of temporal stride during frame sampling.}
We study the effect of temporal stride by training \modelname{} with $\Delta t=0, 8, 15, 30, 45$ on BDD100K, evaluated using linear readout (Figure~\ref{fig:delta_t-sweep}).
Performance peaks at $\Delta t=15$ and degrades only slightly at $8$ and, $30$ while dropping further at values $0$ and $45$.
When it is small, there is limited variance in object appearance, diminishing the value of video data, and when it is too large, correspondence between frames decreases and optical flow becomes unreliable.
Note that for $\Delta t=0$, we add jitter to the initial large crop by up to $10\%$ of the image size.
Figure~\ref{fig:delta_t-vis} shows frame sequences from 2 different videos, highlighting the high variability of motion in BDD100K. 

\subsection{Subcrops as pseudo-iconic training images}
\label{sec:subcrop-analysis}

\begin{figure}[h!]
    \centering
    \hfill
    \begin{subfigure}[b]{0.45\textwidth}
        \centering
        \includegraphics[width=0.78\textwidth]{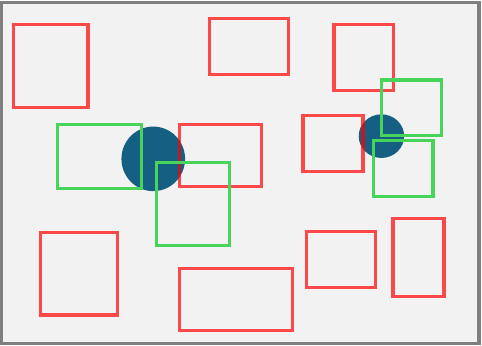}
        \vspace{0.08in}
        \caption{Toy simulation of the probability of subcrop acting hitting a circular object. Subcrop hits are green and non-hits are red. 
        }
        \label{fig:toy-subcrops}
    \end{subfigure}
    \hfill
    \begin{subfigure}[b]{0.45\textwidth}
        \centering
        \includegraphics[clip,trim=0.3cm 0.25cm 0.3cm 0.3cm,width=0.8\textwidth]{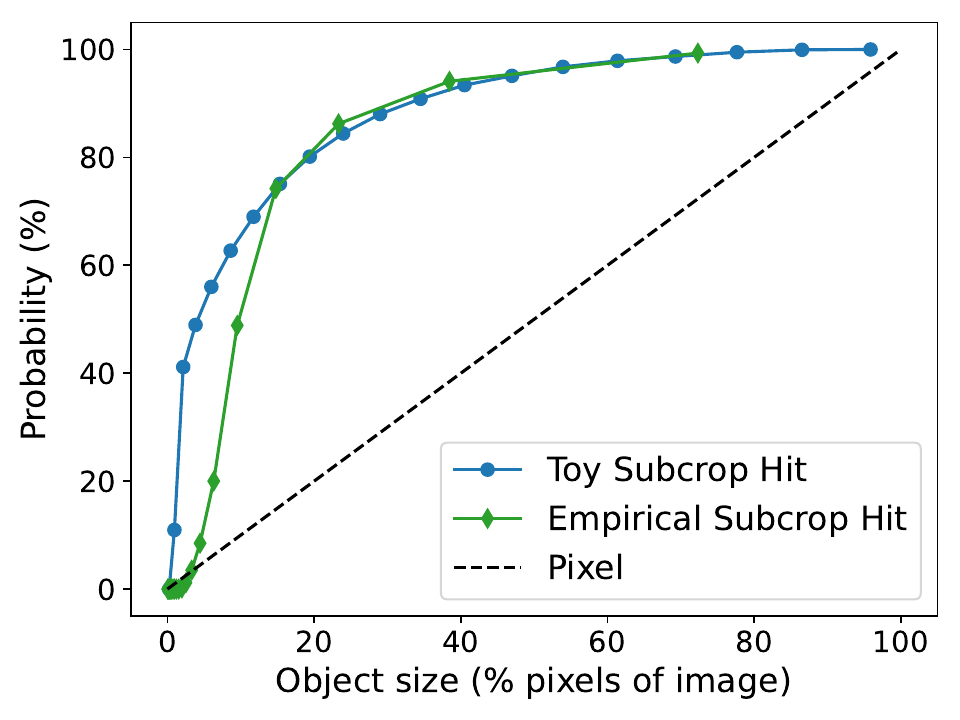}
        \caption{Subcrop and pixel probabilities for foreground objects of varying size from toy and empirical simulation. 
        }
        \label{fig:subcrop-analysis-graph}
    \end{subfigure}
    \hfill
    \caption{Analysis of subcrops as pseudo-iconic views.}
    \label{fig:subcrop-analysis}
\end{figure}

To understand the impact of pseudo-iconic subcrops on \modelname{}'s performance, particularly on small objects, we analyze their effect on object prevalence in the pooled objective.
Using a simulated circular object (Figure~\ref{fig:toy-subcrops}), we calculate the probability of a subcrop capturing it as a pseudo-iconic view, i.e. subcrop ``hit'', and compare this to the probability of a pixel landing on the object, emulating a pixel-level dense SSL objective.
We count a subcrop hit if it reaches at least $5\%$ object coverage, which we justify as background classes generally have little visual variation and consequently, minimal impact on the pooled representations.
We extend this analysis to the BDD100K semantic segmentation dataset, empirically simulating subcrops, and computing subcrop hit and pixel probabilities for varying object sizes. 
The simulation results, illustrated in Figure~\ref{fig:subcrop-analysis-graph}, show a greater relative difference between subcrop hit and pixel probabilities for smaller objects, indicating pseudo-iconic subcrops increase their prevalence in the pooled objective.
This likely contributes towards \modelname{}'s improved performance on small object classes compared to dense SSL methods.
Further details of the analysis can be found in Appendix~\ref{app:subcrop_analysis}.

\section{Conclusion}
\label{sec:conclusion}
Self-supervised learning on naturalistic videos presents many unsolved challenges, especially due to the presence of high-resolution, multi-object crowded scenes with severe spatial imbalance.
Iconic methods rely on single-subject images, and dense methods struggle with the scale imbalance of objects. 
We propose \modelname{} that combines pooled region-invariance learning and dense flow-equivariance learning objectives in a unified framework.
\modelname{} achieves state-of-the-art performance on downstream semantic segmentation and object detection evaluations compared to prior methods pretrained on the same video datasets, particularly on recognizing small objects.
Our study on the effects of crop area, input resolution, and temporal stride also offers key insights on the design choices for video self-supervised learning.

\newpage
\section*{Acknowledgements}
\looseness=-10000
We thank Jenny Zhu for her assistance in generating semantic segmentation labels for the WT-Sem dataset. The work is supported in part by the Institute of Information \& Communications Technology Planning \& Evaluation (IITP) under grant RS-2024-00469482, funded by the Ministry of Science and ICT (MSIT) of the Republic of Korea in connection with the Global AI Frontier Lab International Collaborative Research. AW is supported by the NSERC PGS-D Scholarship. CH is supported by the DoD NDSEG Fellowship. The compute is supported in part through the Microsoft Accelerating Foundation Model Research (AFMR) program, a Google Cloud Platform (GCP) award, and NYU IT's High Performance Computing resources, services, and staff expertise.

\bibliography{ref}
\bibliographystyle{iclr2025_conference}

\newpage
\appendix
\section*{Appendix}

\section{Implementation details}
\label{app:implementation}
\paragraph{Backbone.}
As discussed in the pretraining details, we use a Resnet-50~\citep{he2016deepresnet} as our backbone architecture.
The projector model is a non-linear, 2-layer MLP (linear for pooled, $1\times 1$ convolutions for dense) that has a $4096$ hidden dimension and projects down to $256$ dimensions.
The predictor is the same network with $256-4096-256$ channels.
We follow BYOL~\citep{grill2020byol} with a momentum starting at $0.996$ and increasing to $1$ throughout training.

\paragraph{Decoder details.}
The decoder uses a single Bottleneck block from the ResNet architecture with a $8\times$ downsampling ratio in the number of channels.
Upsampling in the decoder is $2\times$ and the lateral connection is a single linear convolutional layer that up-projects the input latent to match the decoder channels ($1024\rightarrow 2048$ in the first decoder block and $512\rightarrow 2048$ for the second block).
As mentioned, 2 decoder blocks are used to achieve a total of $4\times$ upsampling.

\paragraph{Supervised and self-supervised flow prediction.}
Flow is predicted using a supervised off-the-shelf RAFT model or an unsupervised UFlow~\citep{jonschkowski2020uflow} model that we train ourselves.
For unsupervised training, we exactly follow UFlow and train on the KITTI~\citep{geiger2013kitti} dataset before finetuning on BDD100k~\citep{yu2020bdd100k} for 100,000 steps on daytime-only videos.
The training and inference resolutions were set to $256\times 512$ to better match the inference setting.
KITTI used adjacent frames (10Hz video) while BDD frames were sampled with a temporal stride of 10 (30Hz video).

\paragraph{Local cropping details.} $K=6$ paired local crops are sampled using the methods described. Cropping is performed using RandomResizedCrop with an output resolution of $192\times 192$. Jitter is 10\% of the input image size and a standard aspect ratio range of $[3/4, 4/3]$ is used.

\paragraph{Loss details.}
We sum our 2 loss functions directly and give them equal weight.
The loss computation and warping function were applied to representations after reversing the affine transform and resizing to the input image resolution. This is to take full advantage of high resolution flow like in FlowE~\citep{xiong2021selfflowe}.
We also use flow-based occlusion to prevent misaligning occluded regions without correspondence.
We use the same occlusion formulation as DDFlow~\citep{liu2019ddflow} and parameters $\alpha_1=0.1,\alpha_2=0.5$.
We also mask out regions that are not visible after affine transformations for $\mathcal{L}_\text{dense}$.

Our loss is symmetrical: we reverse the $x_t$ and $x_{t+\Delta}$ so that both are encoded by the online weights and used for optimization at each training step.

\paragraph{Optimization details.}
AdamW is used as the optimizer and a weight decay value of $0.01$.
A learning rate of $5e-4$ is used with 32 GPUs and $4$ image pairs per GPU for a batch size total of $128$.
Cosine learning rate decay is used with a schedule for 300 epochs, despite early termination due to compute limitations.
LR warmup is used for 2 training epochs.
Full \texttt{float32} precision is used during training.

\paragraph{Evaluation settings.}
For all BDD and Cityscapes semantic segmentation and object detection readout tasks, we follow the setup described in FlowE~\citep{xiong2021selfflowe} for ResNet-based methods.
For ViT-based methods, we adopt those settings, but use AdamW for the optimizer with a learning rate of $3e-5$ and weight decay of $0.05$, and a crop size of $512 \times 512$ rather than the normal $512 \times 1024$ to accommodate the square aspect ratio used in ViT pretraining, following the semantic segmentation linear readout setup described in iBOT~\citep{zhou2021ibot}.
In addition, ViT-based methods require sliding window inference in order to achieve performance that is competitive with convolution-based methods.

For ADE20K and WT-Sem linear readout, we simply use the respective BDD linear readout settings for ResNet and ViT methods.
For ADE20K and WT-Sem UperNet finetuning, we follow the procedure described in iBOT~\citep{zhou2021ibot} except we use a batch size of 4 for WT-Sem finetuning.

\paragraph{Ablation data sampling.}
For all ablation experiments, we employ repeated sampling like in MAE-st~\citep{feichtenhofer2022maskedmae-st} which samples $R$ frames each time a video is encountered for faster data loading.
Therefore, each pass through every video in the dataset counts as $R$ epochs.

\section{Compute resources}
The full model is trained on 16 A100s and takes about 30h for 100 epochs on BDD100K or 18min per epoch. Walking Tours takes longer at 40min per epoch, as the number of training samples per epoch is larger.

Ablation-sized experiments were run on 2 or 4 H100/A100 GPUs for a total of 40 epochs, taking 20--40h depending on the configuration.

\section{Subcrop analysis}
\label{app:subcrop_analysis}

For the toy simulation of subcrops, we place a foreground object as a centered circle of varying size within a $256 \times 512$ frame.
We then simulate all possible subcrops of area $A \in [0.02, 0.04, 0.06, 0.08]$.
For each subcrop area, we compute subcrop hits, i.e. whether at least 5\% of the subcrop contains the object, using numerical grid-based integration.
We compute the subcrop hit probability, or subcrop hits over valid subcrops, averaged across subcrop areas, as well as the pixel probability, or object pixels over total image pixels.

We also emulate our training procedure for our empirical simulation of subcrops.
For each of the 7,000 images in the BDD100K semantic segmentation training dataset, we sample two global crops with area sampled from $U[0.16, 0.45]$ and for each global crop, 4096 subcrops with area sampled from $U[0.02, 0.03]$.
We compute subcrop probability and pixel probability independently for the pixels of each foreground class: pole, traffic light, traffic sign, person, rider, car, truck, bus, train, motorcycle, bicycle.
We then group the results into 10\% quantile bins by object size (i.e. pixel proportions) and average the subcrop and pixel probabilities.
We utilize a slightly different subcrop area range in the empirical simulation because our two-step global crop and subcrop procedure results in a logarithmic-like distribution.

\begin{figure}[h!]
    \centering
    \includegraphics[width=0.8\textwidth]{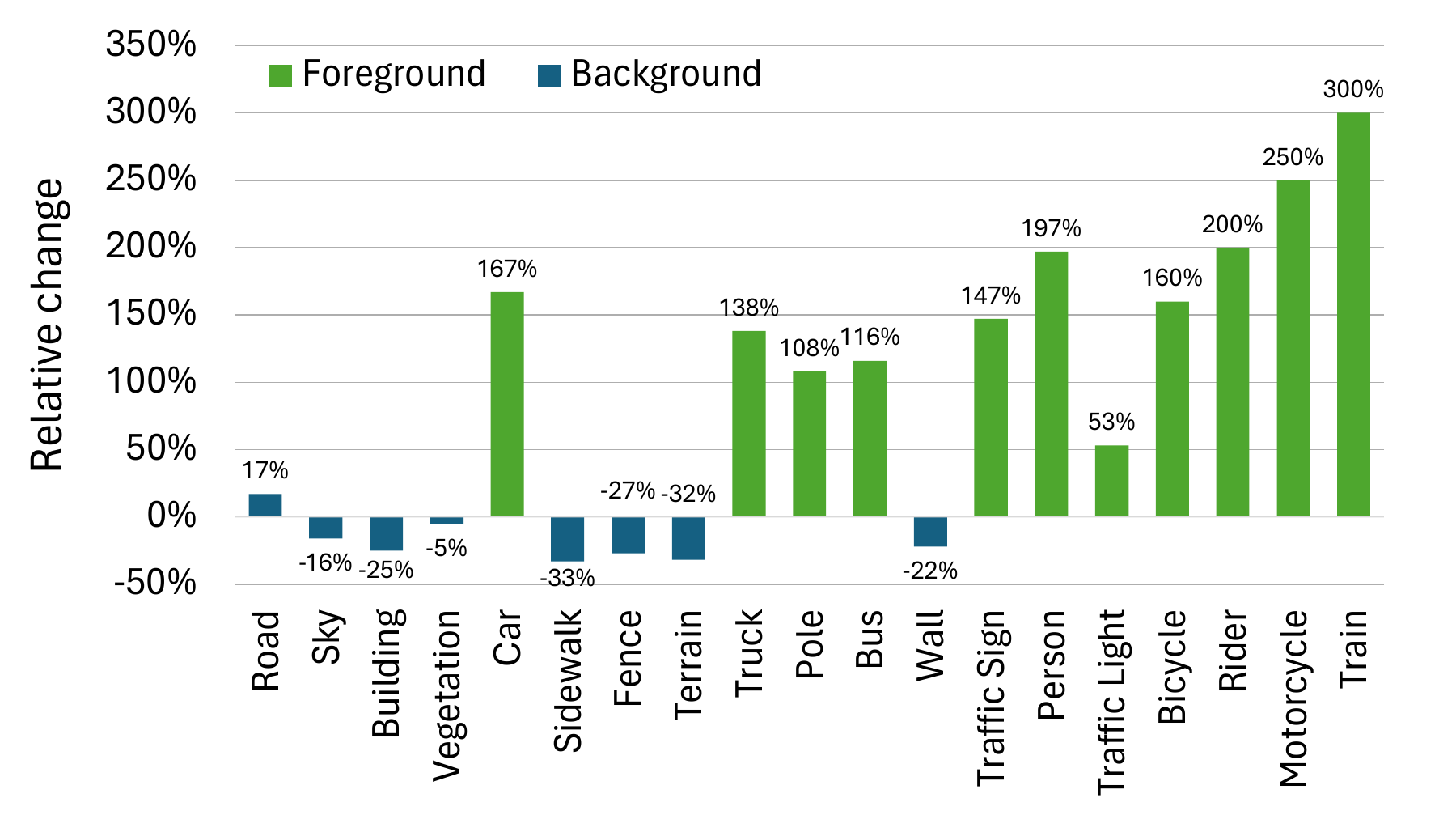}
    \caption{Relative change in local crop region class assignments relative to per-pixel class distribution.}
    \label{fig:relative_class_change}
\end{figure}

We hypothesize that \modelname{}'s improvement on spatially underrepresented classes, as shown in  Table~\ref{table:bdd-class-breakdown}, is due to this subcrop effect.
To quantify this effect on real data, we perform a similar exercise as above on the BDD100K semantic segmentation training set.
We sample subcrops following our method and assign a class label to each subcrop.
If over 10\% of the subcrop is a foreground class (\emph{not} road, sky, building, vegetation, sidewalk, fence, terrain), then we label the subcrop as the majority foreground class.
Otherwise, the majority background class label is assigned.
In Figure~\ref{fig:relative_class_change}, we show the relative change in class distribution when using this subcrop class assignment.
Foreground classes (green) increase in occurrence while background classes (blue) decrease in frequency, besides road.

\newpage
\section{Walking~Tours~Semantic benchmark}
\label{app:wt-sem}

\begin{figure}[h]
\centering
\begin{subfigure}{0.32\textwidth}
    \centering
    \includegraphics[width=\textwidth]{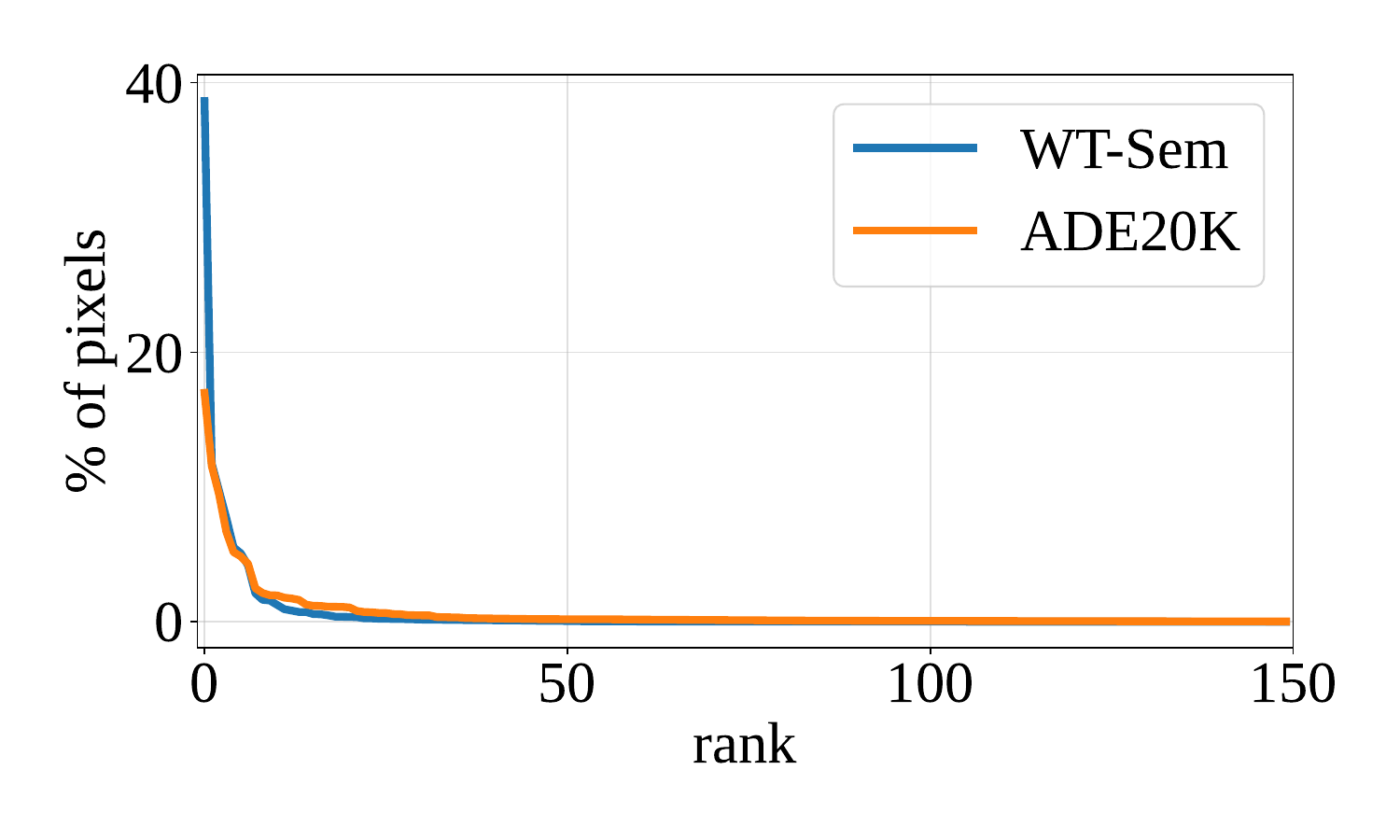}
    \caption{Frequency of classes by pixel}
    \label{fig:wt-sem-pixel}
\end{subfigure}
\hfill
\begin{subfigure}{0.32\textwidth}
    \centering
    \includegraphics[width=\textwidth]{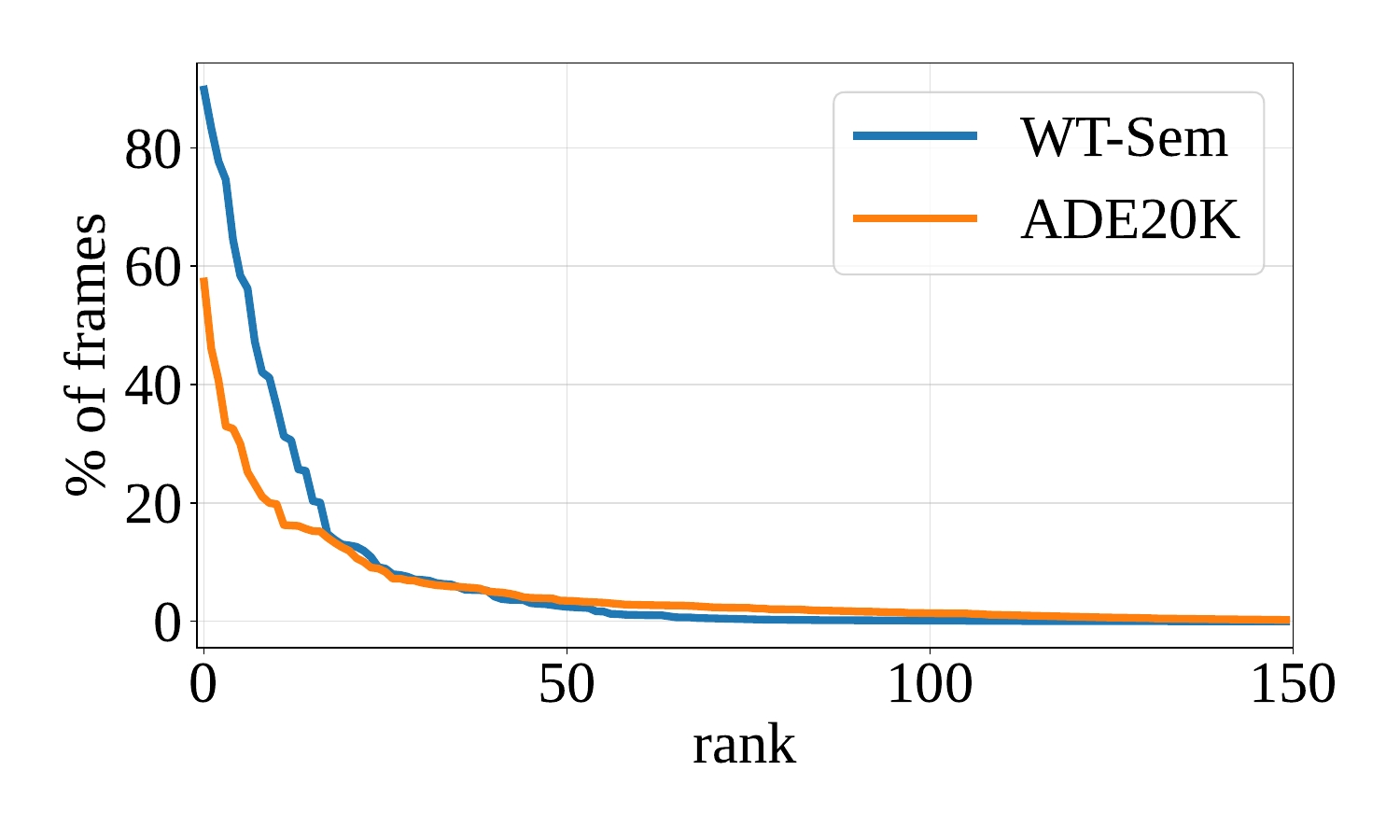}
    \caption{Frequency of classes by frame}
    \label{fig:wt-sem-frame}
\end{subfigure}
\hfill
\begin{subfigure}{0.32\textwidth}
    \centering
    \includegraphics[width=\textwidth]{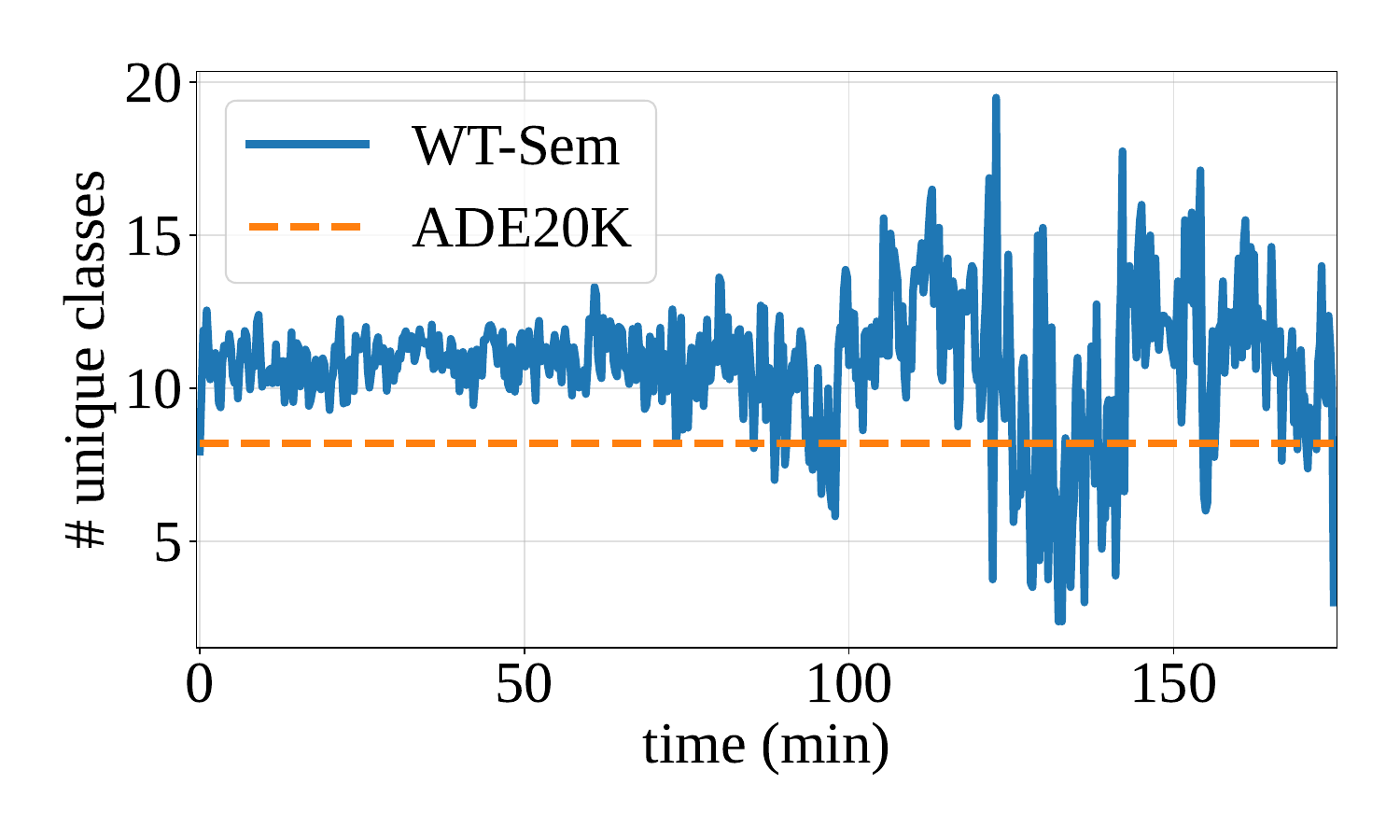}
    \caption{Unique classes over time}
    \label{fig:wt-sem-unique}
\end{subfigure}
\caption{Analysis of WT-Sem in comparison to ADE20K~\citep{zhou2017ade20k} by frequency of each class by pixels occupied (a) or frames present (b) and number of unique classes in each frame (c).}
\label{fig:wt-sem-analysis}
\end{figure}

\begin{figure}[h!]
\centering
\includegraphics[width=0.9\textwidth]{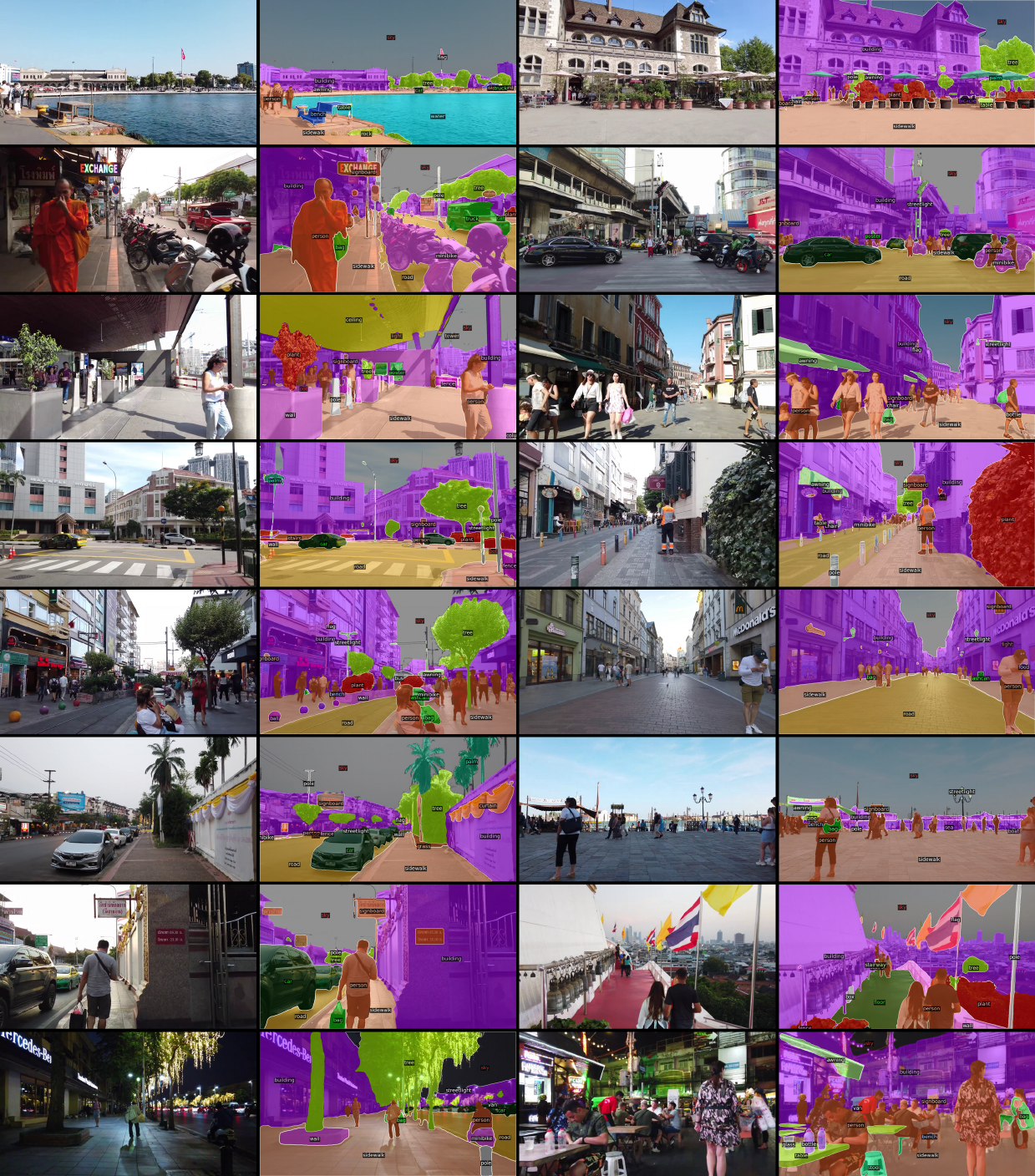}
\caption{Visualizations of images and associated semantic segmentation masks from the WT-Sem benchmark.}
\label{fig:wt-sem-vis}
\end{figure}

We create the WT-Sem benchmark by sampling a frame every 2 seconds from each of the 10 videos in \wtall{} as well as 3 new walkaround videos.
The new walkaround videos are filmed in Rome, Torun, and Poznan, sourced from the same YouTube channel as WT~\citep{venkataramanan2023imagenetdora} under the Creative Commons (CC-BY) license.
The Swin-L variant of OpenSeed~\citep{zhang2023openseed}, pretrained on COCO~\citep{lin2014coco} and Objects365~\citep{Shao2019objects365} and finetuned on ADE20K, is used to generate semantic segmentation masks.
We utilize the 25,910 frames sourced from \wtall{} as the training set and the 6,170 frames sourced from the 3 new videos as the validation set.
Figure~\ref{fig:wt-sem-analysis} shows our analysis of WT-Sem in comparison to ADE20K~\citep{zhou2017ade20k}, where we observe that both datasets have long-tailed class distributions and WT-Sem has slightly higher number of unique classes per frame.
We also visualize examples from the WT-Sem benchmark in Figure~\ref{fig:wt-sem-vis}.

\newpage

\newpage
\section{Additional visualizations}
\label{app:additional-visualizations}
\begin{figure}[h]
\includegraphics[width=0.9\textwidth]{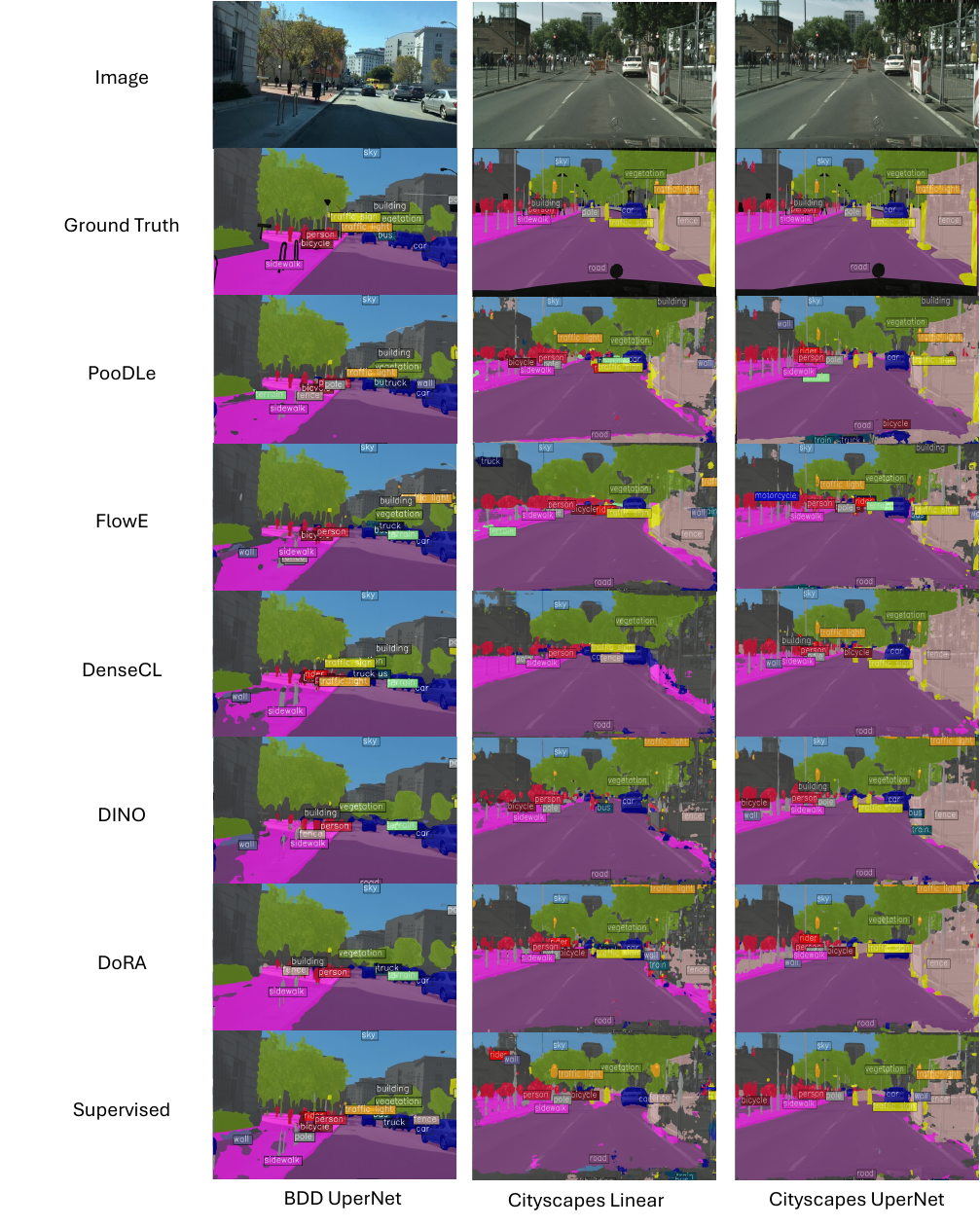}
\vspace{-0.1in}
\caption{Visualizations of semantic segmentation masks for BDD linear readout, Cityscapes linear readout, and Cityscapes UperNet readout.}
\label{fig:big-semseg-vis}
\end{figure}

\begin{figure}[h]
\includegraphics[width=\textwidth]{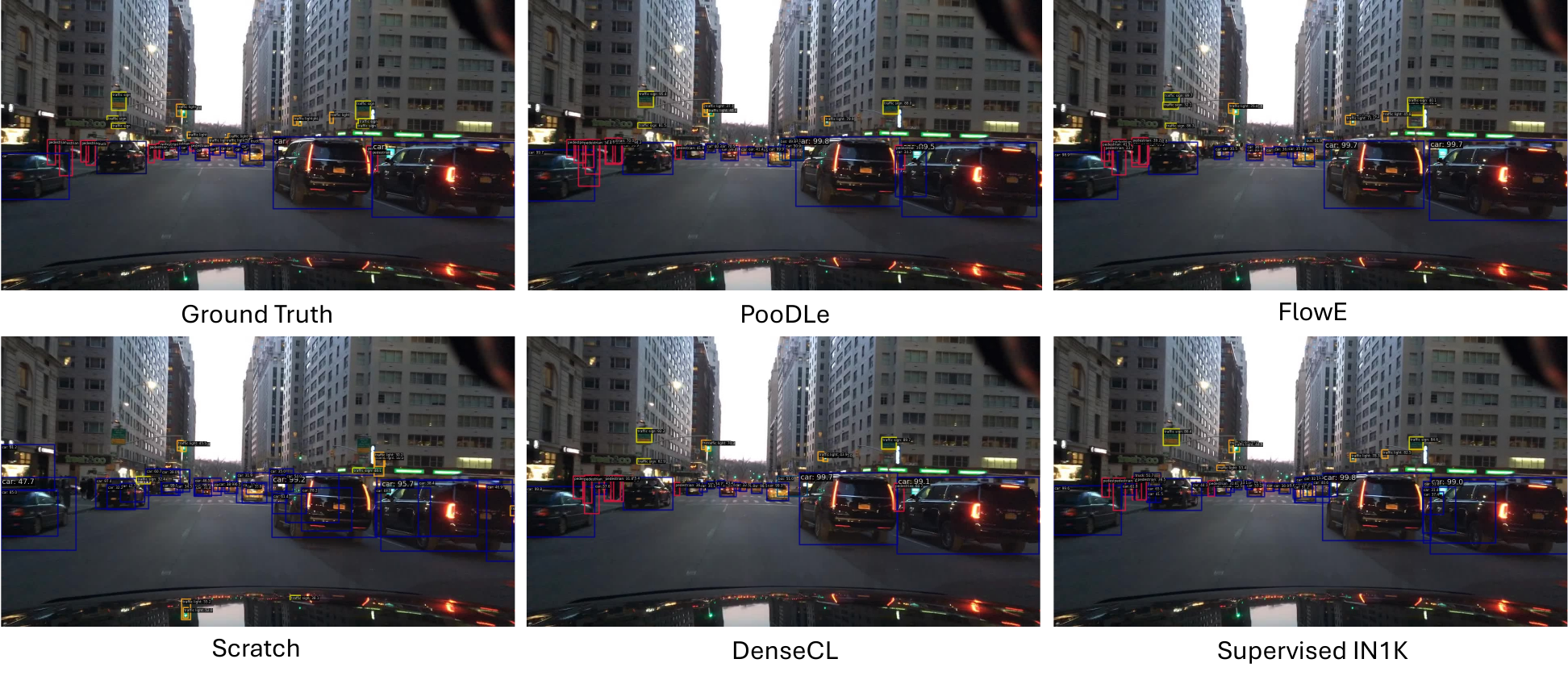}
\vspace{-0.25in}
\caption{Visualizations of object detection bounding boxes for BDD FPN readout. 
}
\vspace{-0.1in}
\label{fig:big-det-vis}
\end{figure}

\begin{figure}[h]
\includegraphics[width=\textwidth]{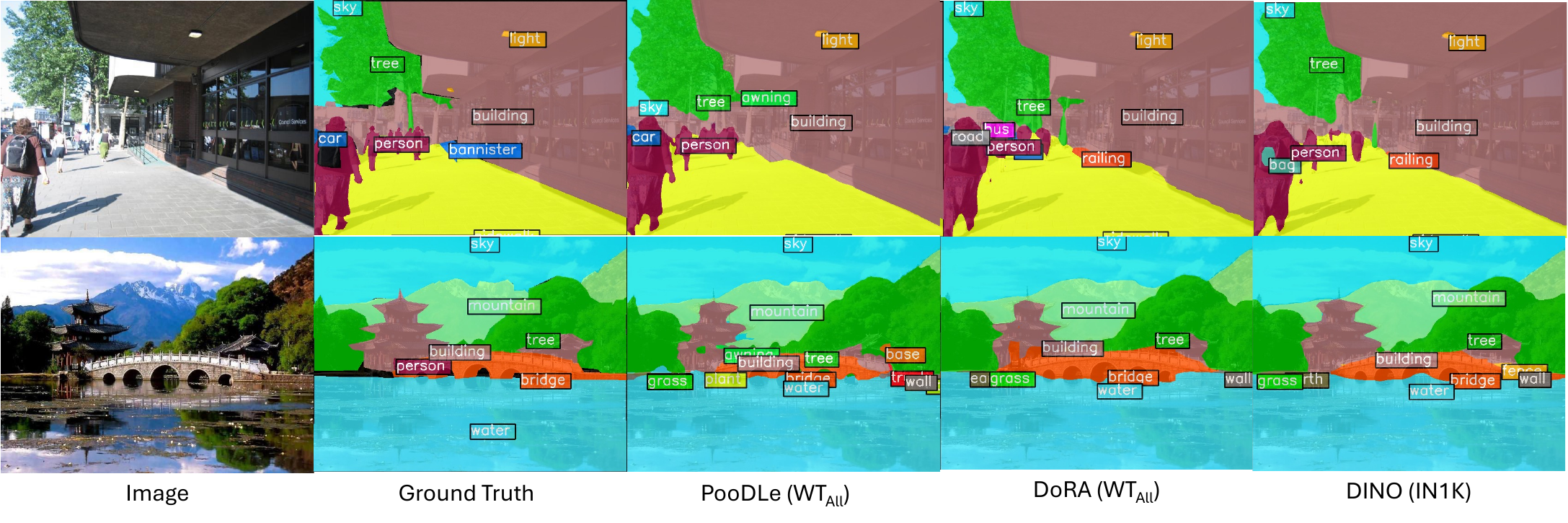}
\vspace{-0.25in}
\caption{Visualizations of semantic segmentation masks for ADE UperNet finetuning.}
\label{fig:big-ade-vis}
\end{figure}

We provide additional visualizations of results on our evaluated benchmarks: BDD100K~\citep{yu2020bdd100k} semantic segmentation (Figure~\ref{fig:big-semseg-vis}), object detection (Figure~\ref{fig:big-det-vis}) and ADE20K semantic segmentation (Figure~\ref{fig:big-ade-vis}).
Once again, we note that \modelname{} produces segmentation maps with clearer boundaries while also effectively capturing small objects.

\newpage

\section{Accuracy values for class breakdown and ablations}
\label{app:acc-for-classbreakdown-ablations}
\vspace{-0.1in}
\begin{table}[h!]
    \centering
        \small
            \caption{Accuracy values for Table 3, class breakdowns}
            \vspace{-0.15in}
        \resizebox{0.6\textwidth}{!}{
\begin{tabular}{ll|r|rr|rr}
\toprule
\textbf{Method} & \textbf{Pretrain} & \textbf{All} & \textbf{Small} & \textbf{Large} & \textbf{Rare} & \textbf{Common} \\ \midrule
DINO & BDD & 86.8 & 12.3 & 88.3 & 2.3 & 87.8 \\
DenseCL & BDD & 84.9 & 2.0 & 86.6 & 0.0 & 86.0 \\
DoRA & BDD & 88.1 & 19.3 & 89.5 & 7.2 & 89.1 \\
FlowE & BDD & 88.5 & 18.2 & 89.9 & 32.0 & 89.2 \\
\hightlightrow
\modelname{} & BDD & \textbf{89.2} & \textbf{33.6} & \textbf{90.3} & \textbf{34.2} & \textbf{89.9} \\ \midrule
Supervised & IN1K & 84.7 & \textbf{36.9} & 85.3 & 23.8 & 85.1 \\
\hightlightrow
\modelname{} & BDD* & \textbf{90.7} & 35.6 & \textbf{91.2} & \textbf{46.9} & \textbf{91.2} \\
\bottomrule
\end{tabular}}

    \label{tab:acc_class_breakdown}
\end{table} 
\begin{table}[h!]
    \centering
    \small
    \vspace{0in}
    \caption{Accuray values for Table 4, ablations }
    \vspace{-0.15in}
        \resizebox{0.98\textwidth}{!}{
        
    \begin{tabular}{l|ccccc|r|rr|rr}
    \toprule
    \textbf{Variant} & \textbf{Dense} & \textbf{Pool} & \textbf{Top-Down} & \textbf{Lateral} & \textbf{Flow}  & \textbf{All} & \textbf{Small} & \textbf{Large} & \textbf{Rare} & \textbf{Common} \\ \midrule
    \textcolor{gray}{1} FlowE & \ding{51} & &  &  & RAFT  & 85.0 & 22.8 & 86.3 & 6.1 & 86.0  \\
    \textcolor{gray}{2} & \ding{51} & \ding{51} &  &  & RAFT & 86.2 & 14.2 & 87.6 & 6.3 & 87.1 \\
    \textcolor{gray}{3} & \ding{51}  & \ding{51} & \ding{51} &   & RAFT & \textbf{86.8} & 11.9 & 87.7 & 13.6 & \textbf{87.7} \\
    \textcolor{gray}{4} & \ding{51} & \ding{51}  &   & \ding{51} & RAFT & 86.6 & 22.1 & \textbf{87.9} & 16.9 & 87.5 \\
    \textcolor{gray}{5} & \ding{51} &  & \ding{51} & \ding{51} & RAFT & 84.2 & 25.5 & 85.3 & 28.2 & 84.9 \\
    \hightlightrow \textcolor{gray}{6}  \modelname{}$\dagger$ & \ding{51}  & \ding{51}  & \ding{51} & \ding{51} & UFlow & 86.0 & 26.4 & 87.2 & \textbf{29.6} & 86.7 \\
    \hightlightrow \textcolor{gray}{7}  \modelname{} &
    \ding{51} & \ding{51} & \ding{51} & \ding{51} & RAFT & 86.5 & \textbf{26.6} & 87.7 & 28.5 & 87.1 \\ \bottomrule
    \end{tabular}}
    \label{tab:my_label}
\end{table}
\newpage
\section{Additional evaluation results}
\label{app:additional-evaluations}

\begin{table}[h]
\caption{Additional BDD semantic segmentation (SemSeg) and object detection (Det) readout evaluations. All settings are conducted with a frozen backbone. $^\ddagger$ BYOL results are taken from FlowE~\citep{xiong2021selfflowe} and used DeepLab~v3~\citep{chen2017deeplabv3} in-place of Upernet~\citep{xiao2018upernet}. *Pretrained on BDD, initialized with supervised ImageNet weights.}
\centering
\small
\resizebox{0.98\textwidth}{!}{
\begin{tabular}{@{}llll|rrrr|rr|rrrr@{}}
\toprule
& & & & \multicolumn{4}{c|}{\bf{BDD100K Sem. Seg.}} & \multicolumn{2}{c|}{\bf BDD100K Obj. Det.} & \multicolumn{4}{c}{\bf{Cityscapes Sem. Seg}} \\
&  &  & & \multicolumn{2}{c}{\bf Linear} & \multicolumn{2}{c|}{\bf UperNet}  & \multicolumn{1}{c}{\bf Det C4} & \multicolumn{1}{c|}{\bf FPN} & \multicolumn{2}{c}{\bf Linear} & \multicolumn{2}{c}{\bf UperNet}  \\
\multirow{-2}{*}{\bf Method} & \multirow{-2}{*}{\bf Arch} & \multirow{-2}{*}{\bf Ep.} & \multirow{-2}{*}{\bf Pretrain} & \multicolumn{1}{c}{mIoU} & \multicolumn{1}{c}{Acc} & \multicolumn{1}{c}{mIoU} & \multicolumn{1}{c|}{Acc} & \multicolumn{1}{c}{mAP} & \multicolumn{1}{c|}{mAP} & \multicolumn{1}{c}{mIoU} & \multicolumn{1}{c}{Acc} & \multicolumn{1}{c}{mIoU} & \multicolumn{1}{c}{Acc} \\ \midrule
Scratch & R50 & - & - & 9.7 & 55.0 & 26.1 & 81.2 & 0.0 & 7.7 & 9.8 & 58.0 & 30.7 & 84.1\\ \midrule
\hightlightrow 
\modelname{} & R50 & 100 & BDD  & 39.2 & 89.2 & 49.9 & 91.8 & 4.9 & 25.2 & 47.2 & 90.2 & 60.7 & 93.5 \\ \midrule
Supervised & R50 & 600 & IN1K & 36.7 & 84.7 & \textbf{55.2} & 92.0 & 3.6 & 24.9 & 46.8 & 87.4 & 63.4 & 93.7 \\
BYOL~\citep{grill2020byol}$^\ddagger$ & R50 & 1000 & IN1K & 28.3 & - & 52.4 & - & 2.8 & 26.0 & 39.9 & - & 60.3 & - \\
DenseCL~\citep{wang2021dense} & R50 & 200 & IN1K & 21.3 & 82.7 & 52.8 & 91.6 & 0.3 & 25.0 & 27.3 & 84.0 & 63.7 & 93.7 \\
Supervised & ViT-S & 300  & IN1K & 41.9 & 88.5 & 50.9 & 91.4 & - & - & 46.8 & 87.4 & 63.4 & 93.7\\
DINO~\citep{caron2021dino} & ViT-S & 800 & IN1K & 38.5 & 88.1 & 52.3 & 92.0 & - & - & 47.1 & 90.3 & 63.6 & 94.0 \\
iBOT~\citep{zhou2021ibot} & ViT-S & 800 & IN1K & 44.4 & 89.6 & 54.2 & 92.2 & - & - & \textbf{52.1} & \textbf{91.5} & \textbf{65.3} & 94.3 \\
\hightlightrow 
\modelname{} & R50 & 100  & BDD* & \textbf{44.7}  & \textbf{90.7} & 54.1  & \textbf{92.7} & \textbf{3.9} & \textbf{28.0} & 52.0 & \textbf{91.5} & 65.1 &	\textbf{94.4} \\ \bottomrule
\end{tabular}
}
\label{table:bdd-additional-results}
\end{table}

We compare \modelname{} against ImageNet-pretrained baselines in Table~\ref{table:bdd-additional-results} and observe that \modelname{} outperforms most baselines except iBOT and ImageNet supervised ViT. 
This result is encouraging, as pretraining on naturalistic video is more challenging due to spatial and class imbalance, yet is also a more realistic setting that enables the use of broader sets of usable data.
Furthermore, we note that pretraining on class-balanced data such as ImageNet particularly benefits mIoU, which weighs all classes equally despite some classes only appearing in a tiny proportion of pixels in evaluation.
Finally, PooDLe pretrained on BDD with weights initialized from the ImageNet supervised checkpoint surpasses all ImageNet-pretrained baselines on linear semantic segmentation.

\section{Per-class evaluation results}
\label{app:per-class-evaluations}

\begin{table}[h]
\centering
\small
\caption{IoU per class on BDD semantic segmentation linear readout. *Pretrained on BDD, initialized with supervised ImageNet weights.
}
\resizebox{\textwidth}{!}{
\begin{tabular}{ll|rrrrrrrrrrrrrrrrrrr}
\toprule
\bf Method & \bf Pretrain & \bf Rd & \bf Sky & \bf Bldg & \bf Veg & \bf Car & \bf Bus & \bf Fence & \bf Truck & \bf Wall & \bf S-walk & \bf Terrain & \bf Train & \bf Pole & \bf Bicycle & \bf Person & \bf M-cycle & \bf Tr. Sign & \bf Rider & \bf Tr. Light \\
\midrule
\bf DINO & BDD & 88.6 & 93.0 & 72.3 & 77.3 & 73.7 & 0.8 & 11.7 & 5.3 & 5.1 & 38.5 & 37.5 & 0 & 8.1 & 0 & 19.6 & 0 & 13.4 & 0 & 17.7 \\
\bf DenseCL & BDD & 82.0 & 88.2 & 68.3 & 72.6 & 63.0 & 0 & 1.5 & 0.5 & 0 & 17.7 & 7.2 & 0 & 0.3 & 0 & 0.1 & 0 & 1.1 & 0 & 9.4 \\
\bf DoRA & BDD & 89.9 & 93.6 & 75.4 & 79.9 & 76.6 & 5.1 & 17.9 & 11.0 & 10.8 & 44.6 & 42.7 & 0 & 13.3 & 0.7 & 25.2 & 0 & 20.5 & 0 & 23.9 \\
\bf FlowE & BDD & 90.6 & 92.9 & 75.8 & 79.6 & 80.8 & 32.9 & 23.5 & 22.7 & 15.3 & 45.7 & 32.4 & 0 & 12.9 & 11.8 & 28.7 & 4.1 & 15.9 & 0 & 12.1 \\
\hightlightrow \bf \modelname{} & BDD & 91.3 & 93.5 & 77.0 & 80.4 & 81.7 & 34.0 & 29.4 & 24.3 & 17.2 & 49.6 & 38.1 & 0 & 24.3 & 18.0 & 35.2 & 2.9 & 26.6 & 0 & 21.2 \\
\midrule
\bf Supv. & IN & 79.8 & 88.8 & 70.0 & 77.2 & 72.6 & 24.9 & 21.8 & 14.4 & 7.2 & 18.4 & 31.8 & 0 & 22.8 & 36.8 & 40.2 & 19.5 & 31.8 & 8.2 & 31.2 \\
\hightlightrow \bf \modelname{} & BDD* & 92.6 & 94.0 & 80.3 & 82.2 & 84.8 & 54.7 & 34.9 & 33.4 & 17.8 & 56.3 & 42.2 & 0 & 25.7 & 27.1 & 41.5 & 7.7 & 39.0 & 0.1 & 35.2 \\
\bottomrule
\end{tabular}
}
\label{table:bdd-per-class-iou}
\end{table}

\begin{table}[h!]
\centering
\small
\caption{Defined groupings and statistics of classes in the BDD semantic segmentation dataset. L=Large, S=Small, C=Common, R=Rare.}
\resizebox{\textwidth}{!}{
\begin{tabular}{l|rrrrrrrrrrrrrrrrrrr}
\toprule
 & \bf Rd & \bf Sky & \bf Bldg & \bf Veg & \bf Car & \bf Bus & \bf Fence & \bf Truck & \bf Wall & \bf S-walk & \bf Terrain & \bf Train & \bf Pole & \bf Bicycle & \bf Person & \bf M-cycle & \bf Tr. Sign & \bf Rider & \bf Tr. Light \\
\midrule
\bf Avg Pix. \% / Im. & 22.0 &18.2 &15.0 &14.4 &8.4 &3.7 &3.4 &3.2 &3.1 &3.1 &2.8 &2.1 &1.0 &0.8 &0.7 &0.6 &0.5 &0.4 &0.4 \\
\bf Total \% of Pix. & 21.3 &17.3 &13.2 &13.2 &8.1 &0.6 &1.0 & 1.0 & 0.5 & 2.0 &1.0 &0.0 &0.9 &0.1 & 0.3 &0.0 & 0.3 & 0.0 & 0.2 \\
\bf Total \% of Im. & 96.5 & 94.8 & 88.4 & 91.7 & 97.3 & 15.0 & 30.6 & 30.5 &15.4 & 66.7 & 36.7 & 0.7 & 95.0 & 6.4 &34.7 & 3.8 & 75.3 & 5.2 & 47.1 \\
\bf Size Grp. & L & L & L & L & L & L & L & L & L & L & L & L & S & S & S & S & S & S & S \\
\bf Freq. Grp. & C & C & C & C & C & R & C & C & R & C & C & R & C & R & C & R & C & R & C \\
\bottomrule
\end{tabular}
}
\label{table:bdd-class-breakdown}
\end{table}

We provide a breakdown of IoU per class on BDD semantic segmentation linear readout in Table~\ref{table:bdd-per-class-iou}.
In Table~\ref{table:bdd-class-breakdown}, we also provide dataset-level statistics for each class computed over the training split of 7,000 images in the BDD semantic segmentation dataset, namely average pixel percentage per image, total percentage of pixels over the dataset and total percentage of images that they appear in over the dataset.
Size and frequency groupings are then independently defined using these statistics and used in Table~\ref{table:bdd-iou-by-class-grouping}.
A class is considered `Large' (L) if its average pixel percentage per image is $>1\%$ and `Small' (S) otherwise.
Separately, we define a class as `Common' (C) if the total percentage of images it appears in is $>20\%$ and `Rare' (R) otherwise.
Notably, \modelname{} achieves significant gains on small classes such as `Pole', `Bicycle', `Traffic Sign', `Traffic Light'. 
Methods trained on BDD underperform supervised IN1K on classes rare in BDD such as `Rider', likely because IN1K offers both abundant and iconic images of these object categories.

\newpage
\section{Backbone computation cost}
We provide a table detailing the number of FLOPs associated with various SSL methods and backbones. We note that our SDM is by far the most efficient upsampling approach for dense representation learning methods.

\begin{table}[h]
  \centering
  \begin{tabular}{lc|c}
    \toprule
    \textbf{Architecture} & \textbf{Associated Methods} & \textbf{GFLOPs} \\
    \midrule
    ResNet-50 & DINO-R50 & 43.3 \\
    ResNet-50 + SDM & PooDLe & 60.5 \\
    ResNet-50 + FPN decoder & PixPro & 124.4 \\
    ResNet-50 + dilated convolutions & FlowE, DINO, DenseCL & 200.7 \\
    ViT-S/16 & DINO, iBOT, DoRA, MAE & 82.9 \\
    \bottomrule
  \end{tabular}
  \caption{Comparison of different backbones, their associated methods, and computational requirements in GFLOPs.}
  \label{tab:architecture_comparison}
\end{table}

\newpage
\section{Flow visualizations}
\label{app:selfsup-flow}
\begin{figure}[h!]
    \centering
    \includegraphics[width=0.7\textwidth]{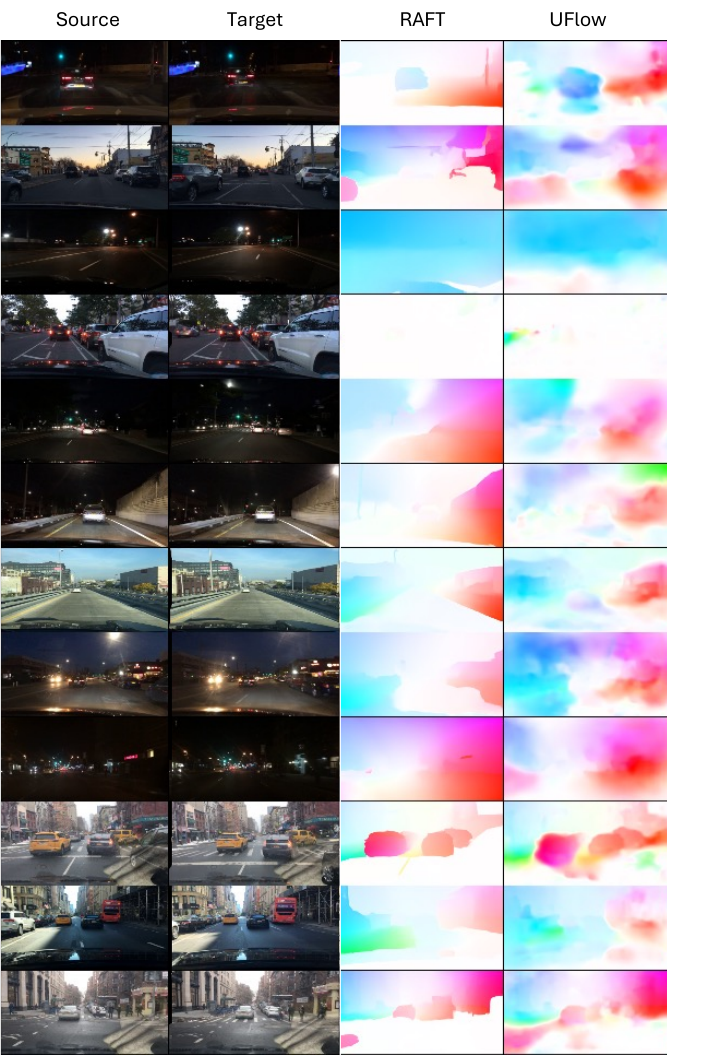}
    \caption{Comparison of predicted optical flow from RAFT (supervised) and UFlow (unsupervised).}
    \label{fig:flow_vis}
\end{figure}

In Figure~\ref{fig:flow_vis}, we compare the predicted flow maps generated from RAFT~\citep{teed2020raft}, an off-the-shelf supervised model, and our own unsupervised UFlow~\citep{jonschkowski2020uflow} model.
The frame pairs are randomly sampled with $\Delta t \in [15, 30]$.
We do note that self-supervised flow, particularly on BDD100K, may exhibit noisy or splotchy results.
This is possibly due to the inconsistent motion and large dark regions that do not offer sufficient photometric supervisory signal. 
This is in contrast to RAFT~\citep{teed2020raft} which learns sharp edges like from supervised labels.
Nevertheless, we find that this self-supervised flow is sufficient for training \modelname{}. 

\end{document}